\documentclass[10pt]{iopart}

\usepackage{soul}
\usepackage{dsfont}
\usepackage{chngpage}
\usepackage{footnote}
\usepackage{multirow}
\usepackage{chngpage}
\usepackage{tabularx}
\usepackage{setspace}
\usepackage{cite}
\usepackage{comment}
\usepackage{graphics}
\usepackage[pdftex]{graphicx}
\usepackage{listings}
\usepackage{algorithm}
\usepackage{algpseudocode}
\usepackage{pifont}
\usepackage{rotating}

\begin{document}

\title{BrainFrame: A node-level heterogeneous accelerator platform for neuron simulations}

\author{Georgios Smaragdos$^1$, Georgios Chatzikonstantis$^3$, Rahul Kukreja$^4$, Harry Sidiropoulos$^3$, Dimitrios Rodopoulos$^5$, Ioannis Sourdis$^2$, Zaid Al-Ars$^4$, Christoforos Kachris$^3$, Dimitrios Soudris$^3$,  Chris I. De Zeeuw$^1$, Christos Strydis$^1$}

\address{$^1$Neuroscience dept., Erasmus MC, Wytemaweg 80, 3015GE, Rotterdam, NL }
\address{$^2$Computer Science and Eng. dept., Chalmers University of Technology, SE-412 96, Gothenburg, SWE}
\address{$^3$MicroLab, National Technical University of Athens (NTUA), 9 Heroon Polytechneiou, 15780, Athens, GR}
\address{$^4$Computer Eng. Lab, Delft University of Technology, Mekelweg 4, 2628CD, Delft, NL}
\address{$^5$imec, Kapeldreef 75, Leuven 3000, BE}
\ead{g.smaragdos@erasmusmc.nl}
\vspace{10pt}
\begin{indented}
\item[]February 2017
\end{indented}

\begin{abstract}
\emph{Objective}: The advent of High-Performance Computing (HPC) in recent years has led to its increasing use in brain study through computational models. The scale and complexity of such models are constantly increasing, leading to challenging computational requirements. Even though modern HPC platforms can often deal with such challenges, the vast diversity of the modeling field does not permit for a homogeneous acceleration platform to effectively address the complete array of modeling requirements.
\emph{Approach}: In this paper we propose and build BrainFrame, a heterogeneous acceleration platform that incorporates three distinct acceleration technologies, an Intel Xeon-Phi CPU, a NVidia GP-GPU and a Maxeler Dataflow Engine. The PyNN software framework is also integrated into the platform. As a challenging proof of concept, we analyze the performance of BrainFrame on different experiment instances of a state-of-the-art neuron model, representing the Inferior-Olivary Nucleus using a biophysically-meaningful, extended Hodgkin-Huxley representation. The model instances take into account not only the neuronal-network dimensions but also different network-connectivity densities, which can drastically affect the workload's performance characteristics.
\emph{Main results}: The combined use of different HPC fabrics demonstrated that BrainFrame is better able to cope with the modeling diversity encountered in realistic experiments while at the same time running on significantly lower energy budgets. Our performance analysis clearly shows that the model directly affects performance and all three technologies are required to cope with all the model use cases.
\emph{Significance}: The BrainFrame framework is designed to transparently configure and select the appropriate back-end accelerator technology for use per simulation run. The PyNN integration provides a familiar bridge to the vast number of models already available. Additionally, it gives a clear roadmap for extending the platform support beyond the proof of concept, with improved usability and directly useful features to the computational-neuroscience community, paving the way for wider adoption.
\end{abstract}

\submitto{\JNE}
%\maketitle
\ioptwocol
%

% ====================================================================
% ====================================================================
\section{Introduction}

% Through the efforts of biologists and computational neuroscientists in recent decades, advance models of cortical neurons were developed using Spiking Neural Networks (SNNs)~\cite{wulfram}. These models do not just abstractly capture aspects of biological processes, like Artificial Neural Networks (ANNs), but directly emulate them. The more complex information encoding in SNNs ensures that they have greater computational capacity~\cite{maass1,maass2} than ANNs and allow neuroscientists to begin making biologically accurate models of brain subsystems, furthering their study. A better understanding of brain functionality can lead to breakthroughs in medicine for tackling brain disease as well as in engineering for inspiring more refined Artificial Intelligence or even new, non-von-Neumann computers. SNNs, thus, are widely used in neuroscientific research to complement biological experiments.

In-vivo and in-vitro experiments are a traditional tool of neuroscientific research. They are powerful experimentation methods, but are also time-consuming and not always reliable. A number of factors can contaminate results like, for example, the influence of anesthesia in in-vivo experiments. What is more, most systemic neuroscientific phenomena require the monitoring of biological systems of very large scale and many such techniques do not allow for this kind of study. Computational neuroscientists use Spiking Neural Networks (SNNs) to cope with such issues. By incorporating SNN models of varied complexity (which themselves are derived by biological experiments) they create predictive simulators that can test their scientific hypotheses and drive more targeted, thus more reliable and refined, biological experimentation~\cite{jornt}.

A major challenge of executing such simulations is the sheer computational complexity that many SNN models entail, compared to simpler modeling classes. Traditional methods of computing, in which the common simulation tool-flows (such as MATLAB or specific neuromodeling tools like NEURON or Brian) are executed, are not up to the task of simulating neural networks of realistic sizes and high detail within a reasonable timeframe for brain research. High-Performance Computing (HPC) has been recently recognized as being able to provide a variety of solutions to cope with this limitation~\cite{chatz, hoang, smaragdos2, glackin2, bhuiyan, Yamazaki_NN_2013}. Unfortunately, the challenge of executing such simulation applications does not stop just at providing the necessary computational power.

In scientific applications such as neuronal simulations, modeling accuracy has a direct impact on simulation speed. The variety of options of viable SNN models used in studies is significant. Every type of model has scientific merit, depending on the subject under study, and models exhibit different characteristics when treated as computational workloads~\cite{bhuiyan, Smaragdos2016}. Modeling features like the inter-neuron connectivity density (the modeling of which also varies according to the biological system under study) can break the embarrassingly parallel (data-flow compatible) nature that most neuron models have, significantly changing the behavior of the application.

Depending on the desired model characteristics, we identify two general types of simulations that are relevant in neuroscientific experiments. The first one has to do with highly accurate (biophysically accurate and even accurate to the molecular level) models of smaller-sized networks that requires \emph{real-time} or close to real-time performance. These kinds of experiments can be used with artificial real-time set-ups or brain-machine interfaces (BMI) and are closely related to brain-rescue studies (TYPE-I experiments). The second type involves the simulation of large- or very large-scale networks in which accuracy can often be relaxed. These experiments attempt to simulate network sizes and connection densities closely resembling their biological counterparts (TYPE-II experiments)~\cite{glackin2}~\cite{Markram}. This, in combination to the variety of models commonly used, makes for a class of applications that vary greatly in terms of workload, while also, depending on the case, requiring high throughput, low latency or both. A single type of HPC fabric, either software- or hardware-based cannot cover all possible use cases with optimal efficiency.

A better approach is to provide scientists with an acceleration platform that has the ability to adjust to the aforementioned variety of workload characteristics. A heterogeneous system that integrates multiple HPC technologies, instead of just one, would be able to provide this. In addition, a framework for a heterogeneous system using a popular user interface for all integrated technologies can also provide the ability to select a different accelerator, depending on availability, cost and performance desired.

Such a hardware back-end must overcome additional challenges to be used in the field. It requires a front-end which should provide two crucial features:

\begin{itemize}
    \item An easy and commonly used interface through which neuroscientists can employ the platform, without the constant mediation of an engineer.
    \item A front-end that can reuse the vast amount of models already available to the community.
\end{itemize}

% Developing and executing experiments with SNN models is a very rigorous process since experimenting with the models presupposes their careful fitting to experimental data. The neuroscientist should be able to interface with the acceleration platform directly, which is not a standard practice today and incurs significant delays in the research process. Lastly, the ability to program the accelerator platform in commonly used coding languages and the portability of legacy code, is essential for wide adoption of the HPC technologies by the community.

In this paper, we propose a framework for an \emph{heterogeneous acceleration platform} for computationally challenging neuroscientific simulations called \emph{BrainFrame}. By using this system, we demonstrate the effect of model characteristics on performance and thus make a concrete case for the significance of employing heterogeneity in HPC systems used the field of computational neuroscience. To this end, we use a state-of-the-art, extended Hodgkin-Huxley (biophysically-meaningful) model~\cite{HH} of the inferior-olivary nucleus (abbrev. InfOli) as a benchmark to evaluate the framework. We chose this model as a respective workload of such neuron representations, as their efficient simulation poses a significant engineering challenge. Even though this model is not the most biophysically accurate representation in the field, it is one of the most accepted and widely used models for brain simulations. We evaluate BrainFrame using three distinct instances of the workload, each differentiated by the presence and complexity of the neuron interconnectivity modeling, leading to vastly different computational requirements, while still reflecting realistic neuroscientific experiments. We propose a front-end for the framework based on the PyNN language~\cite{pynn}. PyNN has been widely adopted by the computational-neuroscience community and has direct integration with many other well-known neuron modeling frameworks, covering both aforementioned features that such a front-end would require.

\section{Methods}

\subsection{The Inferior Olive}

The inferior-olivary nucleus forms an intricate part of the olivocerebellar system, which is one of the most dense brain regions and plays an important role in sensorimotor control. Activity in the inferior olive probably only directly triggers movements, when it is synchronized among multiple neurons~\cite{DeGruijl8937, Hoogland}. In addition, the olivary neurons can provide rhythm and coordination signals for motor functions~\cite{dezeeuw}. It is considered to be imperative for the instinctive learning and smooth completion of motor actions~\cite{gao}. The olive provides one of the two main inputs to the cerebellum through the climbing fibers.

What makes the inferior-olive neurons special is their dense interconnection through electrical connections called gap junctions (GJs), which differ from typical synapses in that they are purely electrical. The gap junctions facilitate the synchronization behavior between the olivary neurons and, subsequently, influence the synchronization and learning properties of the entire olivocerebellar system~\cite{dezeeuw}.

% ====================================================================
\subsection{The InfOli Workload}

\begin{figure}[!t]
    \begin{center}
      \includegraphics[width=0.45\textwidth]{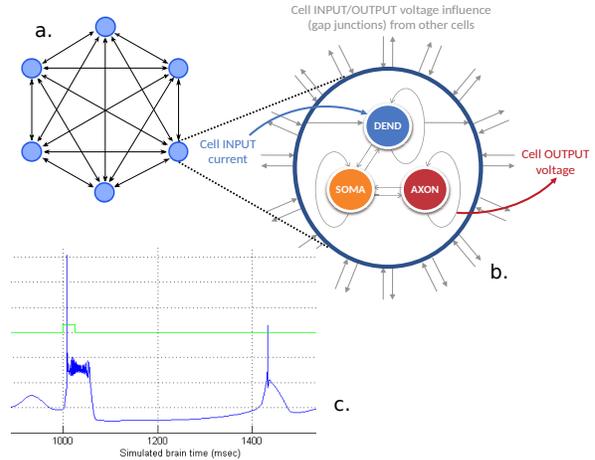}
    \end{center}
    \vspace{-0.2cm}
    \caption{Graphical representation of the inferior-olivary network model. a) 8-neuron network b) single-neuron model in detail c) sample axon response.}
    \label{io-star}
\end{figure}

In this work, a detailed inferior-olive (InfOli) model is considered, which was originally developed by De Gruijl et al.~\cite{jornt}. It implements a neuron with three distinct compartments, the dendrite, the soma and the axon. Within the dendrite, the model also includes gap junctions (thus the characterization as ``extended"), while the cell output represents the input to the climbing fibers (Figure~\ref{io-star}b). The GJs are associated with important aspects of cell behavior as they are not just simple connections; rather, they involve significant and intricate electrical processes, which is reflected in their modeling details.

Every compartment includes a number of state parameters denoting its electrochemical state and the \emph{neuron state} as a whole. The neuron states are updated at each simulation step; every new state update is based upon: The neuron state of the previous simulation step of the executed neuron, the previous dendritic states coming from the GJ connectivity and the externally evoked input to the network, representing the input coming from the rest of the cerebellar circuit.

The three compartments and GJs are evaluated/updated concurrently at each simulation step. The model is calibrated with a simulation time step of $\delta=50~\mu sec$. Simulations steps are identical to each other in terms of operations performed. This simulation step also defines the \emph{real-time} behavior of the whole network. Figure~\ref{io-star}a depicts a representation of the InfOli network model. This network model defines effectively a transient simulator through computing discrete output axon values in time steps which, when integrated in time, recreate the output response of the axon (Figure~\ref{io-star}c).

The InfOli network must be synchronized in order to guarantee the correct exchange of previous dendritic data within a step. Thus, the execution can only be parallelized in space (simultaneous evaluation of neurons within a simulation step), but not in time (parallelization of multiple simulation steps). The cells -- even when not actively spiking -- present an oscillatory behavior, thus affecting network synchronization. As a result, event-driven execution of the network model is not an option.

\begin{table}
    \caption{\label{reqs}Neuron compute requirements per simulation step. }
    \begin{indented}
    \item[]\begin{tabular}{@{}ll}
    \br
    \textbf{Computation} & \textbf{FP Operations / neuron} \\
    \mr
    Gap Junction &12 per connection\\
    Cell Compartment &859\\
    \br
    \textbf{I/O and storage} & \textbf{FP Operations / neuron} \\
    \mr
        	Neuron States				  & 19	             \\
        	Evoked Input				  & 1	             \\
        	Connectivity Vector 		  & 1 per connection \\
        	Neuron Conductances			  & 20               \\
        	Axon Output				      & 1 (Axon Voltage) \\
    \br
    \textbf{Neuron Computation Task} & \textbf{\% of FP ops for 96 cells} \\
    \mr
        	Compartmental Computations	  & 43	             \\
        	Gap Junctions			      & 57               \\

    \br
    Computations per step: & $859*N + 12*N^2*C$\\
                           & C: Connectivity Density \\
                           & N: Network size  \\
    \br
    \end{tabular}
    \end{indented}
    \vspace{-0.2cm}
\end{table}

By profiling the application using a operation and memory-access profiler~\cite{profiler}, it is revealed that the GJs have great impact on the total model complexity. As seen in Table~\ref{reqs}, the total number of floating-point (FP) operations needed for simulating a single step of a single cell including a single GJ are 871\footnote{Table numbers have been updated to amend a profiling mistake reported in previous work~\cite{Smaragdos2016}.}. For many complex experiments, it is not the number of connections but, rather, the \emph{connectivity density} (C) that is indicative of neuron interconnectivity. That is, the \emph{average percentage of the total neuron inventory to which neuron cells are connected} (measured in \%), whereby the complexity becomes quadratic. This makes GJ computations the prevalent contributor, as they break the dataflow nature of the application and dominate computational demands. This is true even for small-scale networks. As an example, for a 96-cell, all-to-all connected network (Table~\ref{reqs}) the GJs comprise almost 60\% of the overall computations.

% ====================================================================
\subsection{Application Use Cases}
\label{subsec:cases}

For our analysis, we employ three use cases drawn from~\cite{Smaragdos2016}, which are representative of the memory and computational requirements of the InfOli workload. All of the use cases are realistic instances of the InfOli application and have neuroscientific merit. They can also be considered as plausible instances of multi-compartmental modeling using HH models with various cases of modeled inter-neuron connectivity.

The application allows for the connectivity of the InfOli network to be programmable by the user before the simulation is deployed. Network connectivity (when present) is defined by an $N \times N$ \emph{connectivity matrix} (where $N$: Network size) of FP weights signifying the weight of each connection. Weights are used in the GJ computations to calculate the connection impact on each neuron. The three use cases focus around the biological complexity of the GJs:

\begin{enumerate}
\item \textbf{InfOli with Realistic Gap Junctions (RGJ)} -- InfOli HH cells modeled with (biophysically) realistic GJ kinematics as presented in~\cite{jornt}. The highest amount of detail is included in the GJ modeling.
\item \textbf{InfOli with Simplified Gap Junctions (SGJ)} -- InfOli HH cells modeled with simplified GJ kinematics. This constitutes a simpler connectivity compared to the RGJ use case.
\item \textbf{InfOli with No Gap Junctions (NGJ)} -- InfOli HH cells modeled without GJ kinematics modeling. This is the simplest use case, whereby the neurons are modeled as independent computational islands.
\end{enumerate}

\begin{figure}
    \centering
    \includegraphics[width=0.5\textwidth]{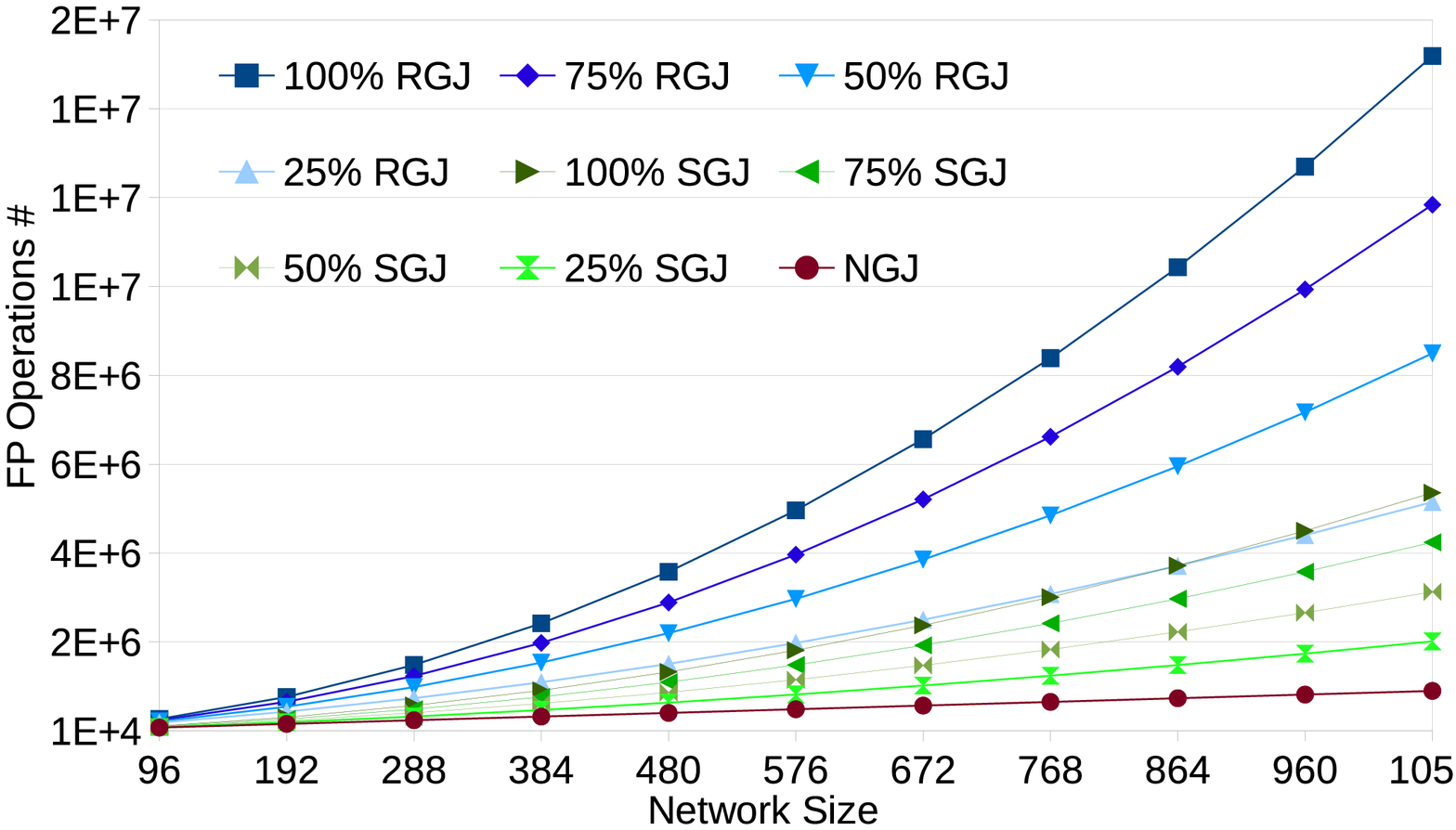}
    \caption{Floating-point operations required per simulation step of the InfOli model for each use case and for different connectivity density percentages (\%).}
    \label{fp_ops}

    \includegraphics[width=0.5\textwidth]{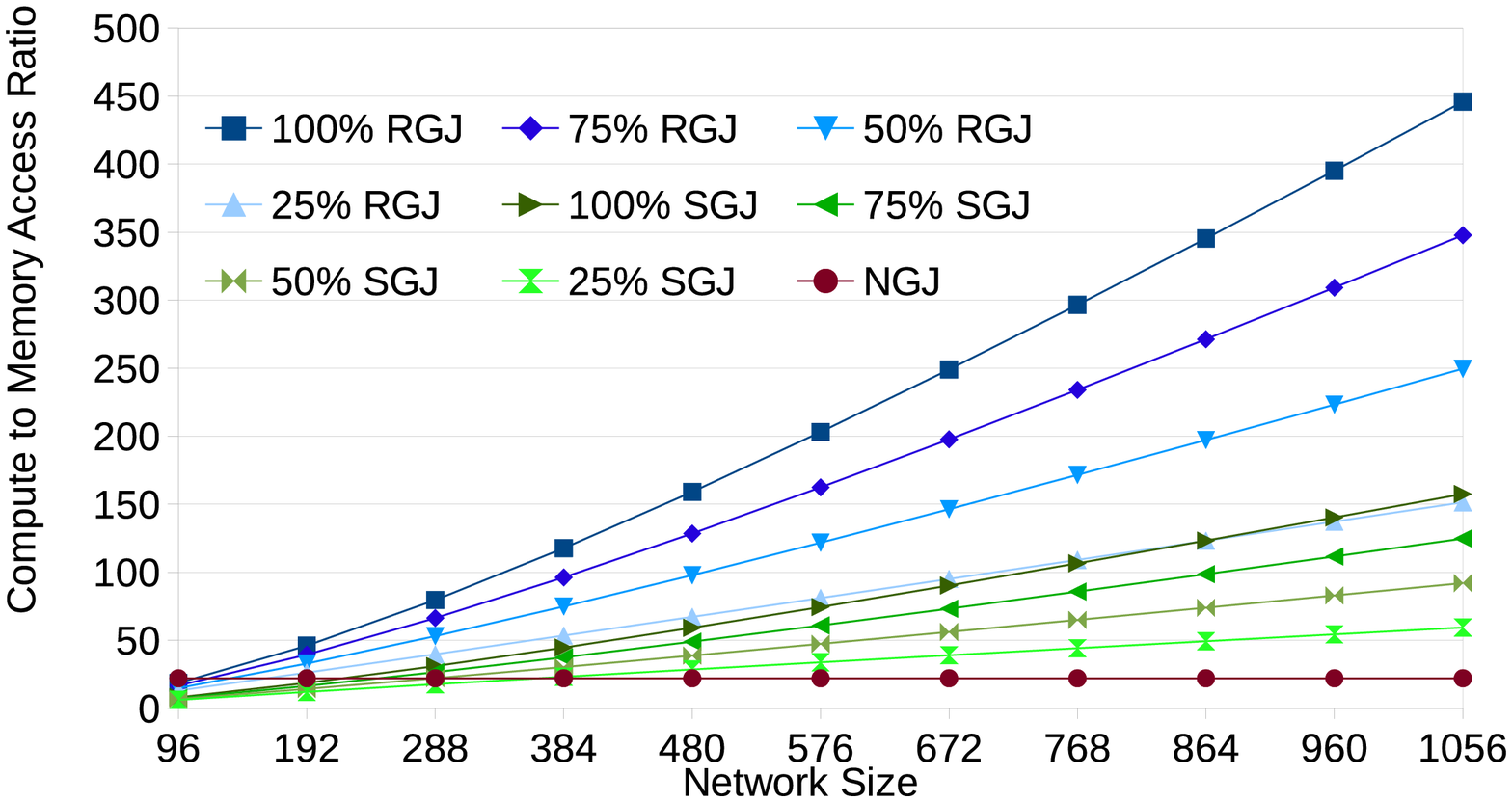}
    \caption{Compute-to-Memory-Access Ratio per simulation step of the InfOli model for each use case and for connectivity density percentages (\%).}
    \label{MemCom}
\end{figure}

In Figure~\ref{fp_ops}, we see the amount of FP operations, based on the aforementioned profiling of the InfOli application. The FP operations are calculated for each of the aforementioned use cases for different connectivity densities. From the same profiling run we can derive the compute (in FLOPS) to memory (in single-FP memory accesses) ratio for the application, that reveals whether each use case is computation- or memory-bound (Figure~\ref{MemCom}).

% ====================================================================
\subsubsection{InfOli with Realistic Gap Junctions (RGJ)}

This use case represents a fully featured version of the InfOli application. The complex Gap Junction dominates the computation in this use case. GJs here are implemented as a very specific representation of the biological nucleus (Algorithm~\ref{gj}). Each cell $C$ in a population of $N$ cells accumulates the influence of an interconnected cell to it (through a GJ) by subtracting its own dendritic voltage (\texttt{prevVdend}) from the dendritic voltage of that cell (\texttt{neighVdend[i]}). It, then, accumulates the resulting voltage influence in an aggregate current \texttt{Ic}, by factoring in the respective GJ-connection weight (\texttt{C[i]}).

\begin{algorithm}[t!]
    \caption{Example of RGJ implementation in C}
    \label{CHalgorithm}
    \begin{algorithmic}[1]
        \setstretch{1.25}
        \For{i=0; i$<$InfOli\_N\_INPUT; i++ }
        \State  V = prevVdend - neighVdend[i];
        \State f = 0.8*V*exp(-1 * V * V/100) + 0.2;
        \State Ic = Ic + (C[i] * f * V);

        \EndFor
        \State return Ic;
    \end{algorithmic}
    \label{gj}
\end{algorithm}

The compute-to-memory-access ratio (from Fi\-gure~\ref{MemCom}) suggests also that this use case is strongly computation-bound for all connectivity cases: With increasing problem sizes, the computations increase at a much faster pace than the memory-access requirements.

%\begin{lstlisting}[frame=single,label=gj,basicstyle=\small, caption=Example of RGJ implementation in C.]
%  for (i=0; i<InfOli_N_INPUT; i++) {
%     V = prevVdend - neighVdend[i];
%     f = 0.8*V*exp(-1 * V * V/100) + 0.2;
%     Ic = Ic + (C[i] * f * V);
%  }

%  return Ic ;
%\end{lstlisting}

% ====================================================================
\subsubsection{InfOli with Simplified Gap Junctions (SGJ)}

The level of detail as in the RGJ case is useful for many modeling experiments but is also an overkill in many other cases that more simple rudimentary connection are involved (like simple synapses that accumulate inputs). Lighter workloads are represented by the SGJ case. We assume a use case of the InfOli application that simplifies the connection between neurons to a few simple input accumulators. The computations per simulation step are now $859*N+4*N^2*C$. The accumulation is parameterized using the weight that is assigned to each connection between two neurons, thus the connectivity information needs to be accessed the same way as is in the RGJ case. The actual FP operations are reduced by about one order of magnitude compared to the previous use case (see Figure~\ref{fp_ops}). Yet, the connectivity aspect still disrupts the pure dataflow nature of the model. A high compute-to-memory ratio is seen here as well, since the computations still increase at a faster pace than the memory requirements.

% ====================================================================
\subsubsection{InfOli with No Gap Junctions (NGJ)}

This is the case where the application becomes purely dataflow and can achieve the greatest parallelism possible. The processing requirements scale almost linearly with the network size and, compared to the other use cases, fewer computations are needed, as shown in Figure~\ref{fp_ops} (computations per simulation step : $859*N$). As we can see in Figure~\ref{MemCom}, although the NGJ use case shows that computation is still the most important aspect of the application, both computation and memory access scale linearly and at a similar pace.
% ====================================================================
\subsection{HPC Fabrics and Implementation}

\begin{table*}
    \begin{center}
    \caption{\label{specs}Specifications of the accelerator fabrics used.}
    %\begin{indented}
        \begin{small}
            \item[]\begin{tabular}{@{}llll@{}}
            \br
            \textbf{Specification} & \textbf{Maxeler DFE (Maia)} &  \textbf{Intel Xeon Phi CPU (5110P)} &  \textbf{NVidia GPU (Titan X)}\\
            \br
            On-Board DRAM      	    & 48 Gb			                 & 8 Gb  			& 12 Gb \\
            RAM bandwidth		    & 76.8 GB/s 		             & 320 GB/s 		& 336.5 GB/s \\
            Memory streams/channels	& 15 			                 & 16 			    & --	\\
            On-chip memory 		    & 6 MB (FPGA BRAMs)              & 30 MB (L2 cache) & 3 MB (L2 cache)\\
            Number of chip cores    & -- 	                         & 61 			    & 3072 CUDA Cores \\
            Chip frequency		    & Depends on design kernel       & 1.053 GHz 		& 1 GHz \\
            Instructions set	    & n/a 		                     & 64 bit 		    & 32 bit \\
            Power consumption (TDP)	& 140 W 		                 & 225 W 		    & 250 W \\
            IC process			& 	65nm  &		22nm	    & 28nm	\\
            \br
            \end{tabular}
            %\end{indented}
        \end{small}
    \end{center}
    \vspace{-0.4cm}
\end{table*}

Our heterogeneous platform incorporates three accelerator fabrics; a Maxeler \emph{Maia} Data-Flow Engine (DFE) board~\cite{Pell2013}, an Intel Xeon Phi 5110P CPU~\cite{jeffers2013intel} and a Maxwell-based Titan X GPU by NVidia~\cite{nvidia} (Table~\ref{specs}). All there boards are PCIe-based which is how they communicate with the host system. The use of PCIe interfaces ensures that composition of BrainFrame-enabled machines can been easily tailored on a per-case basis depending on the availability of funds and hardware resources of a research laboratory. Different types and mixes of PCIe-based accelerators can be selected.

The Maia DFE is a Maxeler HPC technology based on reconfigurable hardware. Its tool flow is designed and optimized to accommodate the acceleration of dataflow applications; that is, applications with the bulk of their implementation using purely raw computations with the absence (partially or totally) of branching execution or feedback paths. The Maxeler tools can exploit the nature of dataflow applications to implement uniquely massive pipelines, maximizing the throughput and overall performance. The DFE boards also incorporate a high-bandwidth, multichannel, highly parallel, customizable interface to the onboard DRAM memory resources (up to 96~GBs) making it ideal for scientific applications. The DFE board used in our experimental setup is a 4th-generation Maia-DFE board implemented using an Altera Stratix V 5SGSD8 chip.

The Xeon Phi is a Many Integrated Core (MIC) architecture co-processor which features 61 cores, each capable of supporting up to 4 instruction streams. The generation of Phi cards used in this work, named Knight's Corner, are programmed using well-known programming tools such as OpenMP and OpenCL. However, and in contrast to GPU mentality, the Phi can also be thought of as an accelerator that can act as a stand-alone processor and even features its own Operating System.  This is expected to increase memory-consistency and cache-coherency delays.

GP-GPUs have also been prominent in the HPC domain and in scientific computing in particular. The Titan X includes 3,072 CUDA micro-cores, which are used to parallelize computation execution, and 12~GB of on-board RAM. GPU implementations also benefit from the generally good adoption of the NVidia CUDA-library open environment that allows porting of applications with similar ease to the Phi OpenMP and OpenCL frameworks. GPUs also come at a relatively lower cost than the other two accelerator types. However, as opposed to the the Xeon Phi, a GPU cannot act as its own host increasing communication delays between host and accelerator during execution.

Lastly, it must be noted that BrainFrame is to be used in scientific research that is very dynamic and fast-paced. The goal is not to over-optimize the different accelerator implementations, but to propose and maintain a balance between the programming effort and optimization needed, resulting in shorter development times for cutting-edge research tools. In real research, such development times should be kept short so as not to delay the scientific process.

% ====================================================================
\subsubsection{Infoli on the Maia DFE}

\begin{figure}[!t]
    \begin{center}
      \includegraphics[width=0.5\textwidth]{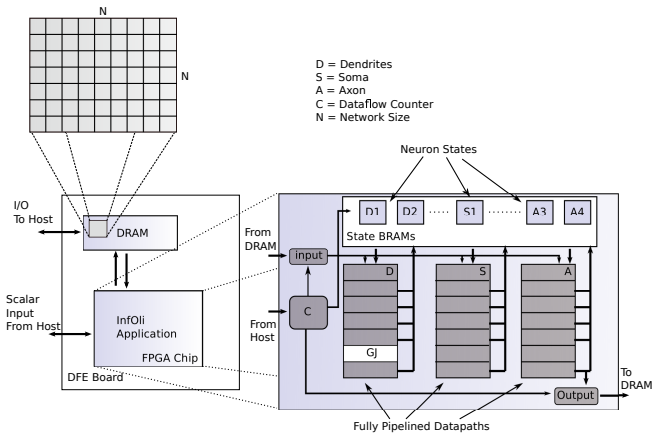}
    \end{center}
    \caption{DFE implementation of the InfOli application.}
    \label{DFEarch}
\end{figure}

The DFE implementation of the InfOli application can be seen in Figure~\ref{DFEarch} and is a more advanced version of the work done in~\cite{smaragdos}. New features include the addition of programmable connectivity and programmable neuron state by the user between experiment runs without the need to re-synthesize the design. The design implements three pipelines on the DFE hardware to accelerate the application, one for each part of a neuron (Dendrite, Soma, Axon), executing the respective computations. The state parameters for each neuron are stored on separate BRAM blocks for fast reading/updating of the network state, as they are the data that are most used throughout the experiment execution. Since every new neuron state is dependent only on the network state of the previous simulation step, a single copy of each neuron state is required at any point during execution. The input stream to the DFE kernel originates in the on-board DRAM and represents the evoked (external) inputs, used in the dendritic computations comprising the network input. The initialization data are also streamed in from the on-board memory only once at the start of execution. The size of the connectivity matrix makes it impossible to store on the on-chip memory. It is, thus, placed on the on-board RAM and streamed in batches dictated by the computations. The kernel output is streamed back to the on-board memory and -- at the same time -- is updated in the (on-chip) BRAM blocks of the DFE.

The program flow is tracked using hardware counters monitoring GJ loop iterations (except for the NGJ case), the neurons executed and the number of simulation steps concluded. The data flows through the DFE pipelines with each kernel execution step (or \emph{tick}) consuming the corresponding input or producing the respective output and new state at the correct execution points according to the hardware counters. DFE execution naturally pipelines the execution of different neurons within one simulation step. Simulation steps are not themselves directly parallelizable, as every neuron must have the previous state of all other neurons available for its GJ computations (only in the RGJ or SGJ cases) before a new step begins. The DFE pipeline is, thus, flushed before a new simulation step begins execution. This dependency is lifted when in the NGJ case.  The GJ calculations form a loop that must finish before the rest of the dendrite-compartment state is calculated. The rest of the dendrite pipeline does not produce valid data for the operation ticks that the GJ influence is being calculated. This delay is partially amortized by using hardware loop unrolling on the GJ calculations, but only to the point that the available chip area allows it. Additionally, in use cases where programmable connectivity is included, the ticks for the evaluation and execution of a GJ connection are always spent regardless of whether a connection actually exists or not. Thus, this implementation cannot benefit from a smaller connectivity density in terms of performance. On the other hand, since one synthesized design can account for all possible connectivity scenarios, the DFE implementation can guarantee predictable performance under all use cases.

% ====================================================================
\subsubsection{Infoli on the Xeon Phi}

\begin{figure}[!t]
    \begin{center}
      \includegraphics[width=0.5\textwidth]{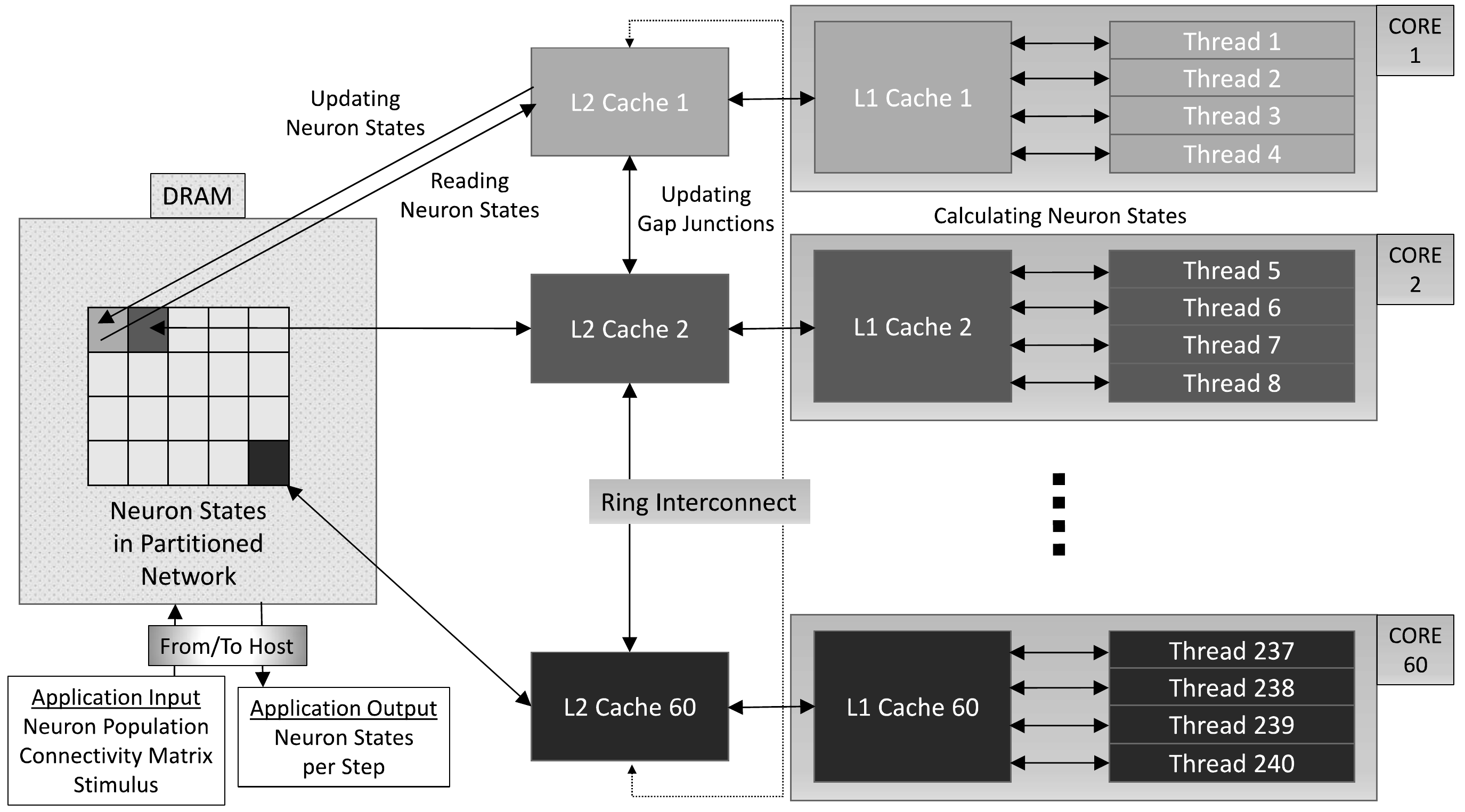}
    \end{center}
    \caption{Xeon-Phi implementation of the InfOli application.}
    \vspace{-0.2cm}
    \label{XEONPhiarch}
\end{figure}

The InfOli application on the Intel Xeon Phi co-processor, depicted in Figure~\ref{XEONPhiarch}, is based on a typical shared-memory implementation. The application uses the OpenMP library to spawn threads, which can work in parallel. As the Xeon Phi 5110P uses one core to handle OS-related tasks and each core features multithreading technology that can service up to 4 instruction streams simultaneously, the InfOli application on the Xeon Phi uses up to $60 \times 4 = 240$ OpenMP threads. Each thread is programmed to handle a part of the neuronal network (sub-network), which is partitioned as uniformly as possible to prevent workload imbalances.

In each simulation step, every OpenMP thread computes its sub-network's state. This process is further broken down into two tasks. Initially, the sub-network needs updated information from the rest of the network, specifically the dendritic-membrane voltage of the other neurons connected with this sub-network (recall Algorithm~\ref{gj}). Thus, each OpenMP thread accesses memory space shared by all threads so as to collect data from other neurons, with the purpose of re-evaluating the state of its sub-network's GJs. In this task, shared-memory accessing can cause stalls in thread operations due to issues such as memory contention.

Upon completion of its first task, each OpenMP thread updates the compartmental states of each neuron in the sub-network. Each of the neuron's three compartments is re-calculated (dendrite, soma and axon). The dendritic compartment specifically uses the updated GJ states evaluated in the previous task in order to assess the incoming current from connected neurons. As already explained, this particular process demands an amount of operations that increases significantly with neuron population in the case of densely connected networks, as we would expect with the increasing computational demands of GJs.

After performing its two tasks for the entirety of its sub-network, each OpenMP thread begins the process anew for the next simulation step, until there are none left. Under this paradigm, the threads operate constantly within the ``timeframe" of the same simulation step. They sync before the execution of a new simulation step, so that stale data from previous simulation steps cannot be exchanged during GJ computation. This behaviour is enforced due to the stiffness of the eHH-model equations, which can be thrown off-balance even by small changes in the numerical data within a single time step. Under a more relaxed model (e.g. a typical HH model), some staleness in data exchange would be more tolerable and the user would be able perform thread synchronization less frequently in order to trade precision for execution speed. Furthermore, it should be noted that the implementation described assumes that the entire network is large enough to be partitioned in $240$ parts. When dealing with smaller networks, the implementation utilizes less than the maximum amount of the platform's assets, since it is designed to require at assign one neuron on each OpenMP thread.

% ====================================================================
\subsubsection{Infoli on the Titan X GPU}

\begin{figure}[!t]
    \begin{center}
      \includegraphics[width=0.45\textwidth]{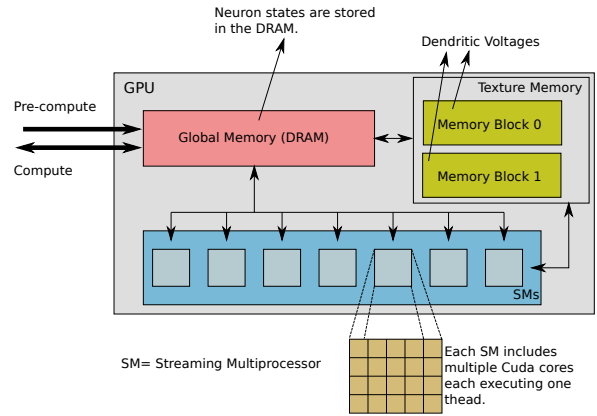}
    \end{center}
    \vspace{-0.2cm}
    \caption{GPU implementation of the InfOli application. Pre-compute and compute operations are issued by the host.}
    \label{gpufig}
    \vspace{-0.2cm}
\end{figure}

In Figure~\ref{gpufig}, we can see the InfOli implementation on the GPU. The execution flow includes two stages, a pre-compute and a compute stage.

In the pre-compute stage, the host initializes the neuron states and the external input currents for the entire simulation duration. It allocates global memory on the device to store the current-step neuron states, next-step neuron states and the external input currents. At the end of this stage, the host copies the required data for simulation onto the GPU. Similarly to the other two accelerator implementations, the current-step dendritic voltages of all cells are accessed frequently as they are used to determine the GJ influence. To reduce memory latency, they are bound to the GPU texture memory. The texture memory is a cached memory on the GPU used to reduce memory latencies when the application has specific memory-access patterns. Writes to texture memory, during the compute stage, are conducted only after all computations of a simulation step have finished. It must be also noted that after the pre-compute stage, no data is transferred from the host to the GPU; the GPU contains all necessary information for the simulation.

\begin{figure*}[!t]
    \begin{center}
      \includegraphics[width=0.8\textwidth]{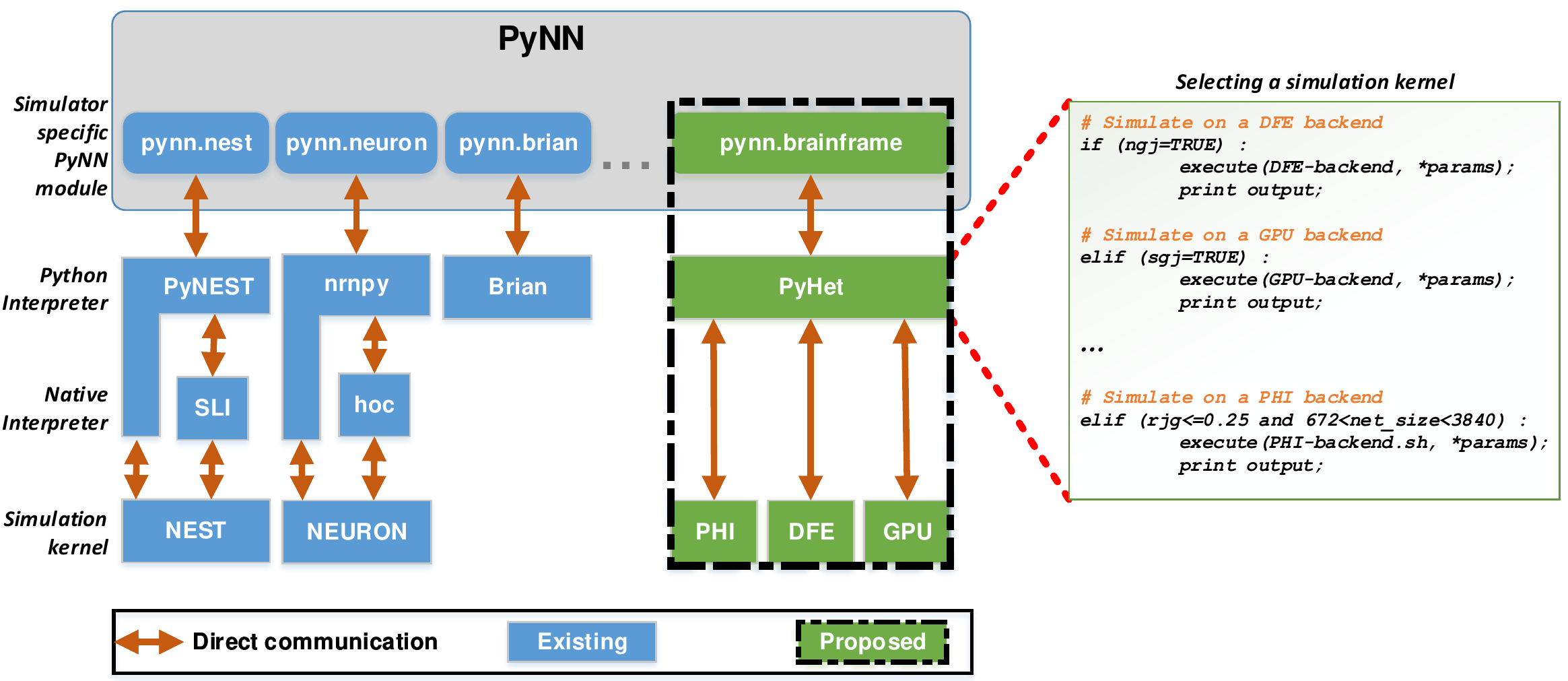}
    \end{center}
    \vspace{-0.4cm}
    \caption{PyNN architecture and the proposed BrainFrame framework.}
    \label{pynn-backend}
\end{figure*}

During the compute stage, the neuron calculations are performed and the new states are persistently stored throughout the simulation duration. To compute the new states for a single simulation step, the host launches a CUDA kernel on the GPU device. Before simulation, the kernel is configured for a particular use case (RGJ, SGJ or NGJ) and inter-neuron connectivity scheme (if applicable). The kernel is executed by a two-dimensional grid of CUDA threads on the device. Threads are executed in parallel by the CUDA micro-cores of the GPU. Every InfOli cell of the model is mapped to a corresponding thread that calculates the states of the neuron. On kernel completion, the host receives the calculated result of the simulation step from the device. The host uses two operation streams to issue the kernel execution and data-transfer operations to the GPU. A kernel in one stream is launched only when the kernel in the other stream has completed. Thus, when one stream is computing the currently executing simulation step, the other stream is performing the necessary data transfers to the host from the GPU. Since the texture memory is updated only after the kernel completes execution, data coherency is maintained. Thus, computation of the current-step neuron states and data transfer of the previously computed states overlap, effectively hiding Host-to-GPU transfer delays.

% ====================================================================
\subsection{BrainFrame \& the PyNN Front-End}
\label{sub:pynn}
PyNN is a Python package that facilitates the interchangeability and the study of different simulation environments within the computational neuroscience community~\cite{pynn}. It allows for simulator-independent specification of neuronal-network models and already supports many popular simulators like NEURON, NEST, PCSIM , Brian, and so on.

The PyNN API supports modeling at multiple levels of abstraction, both at the neuron level and the  network level. It provides a library of standard neuron, synapse and synaptic-plasticity models and a set of commonly-used connectivity algorithms while also supporting custom user-defined connectivity in a simulator-independent fashion.

We integrated the three accelerator fabrics as back-ends on the BrainFrame system using PyNN as a front-end. The PyNN integration provides the neuroscientific community with easy access on the accelerators without constant mediation from the acceleration engineer while also providing an interface for the already established models to be used with the new heterogeneous acceleration back-end. These characteristics of PyNN can have decisive impact on the adoption of BrainFrame by the community.

As a proof of concept for the front-end of the BrainFrame platform, we have added the InfOli model the library of standard PyNN models. Following the PyNN paradigm, the user initially selects the simulator -- in our case our BrainFrame simulator -- and then proceeds to select the neuron model, in our case the Inferior-Olive model. A population of neurons using the chosen model is then generated, determining the inter-neuron connectivity type and, finally, a projection of the specified neuronal network is created.

The main difference between the proposed PyNN-backend substrate and the typical simulator back-ends within the PyNN environment is an additional selection step. In this step, a decision about which of the three alternative acceleration fabrics will be used for a specific experiment is made, based on the available hardware and the characteristics of the simulated neural network.

A conceptional view of the architecture of the PyNN BrainFrame module is shown in Figure~\ref{pynn-backend}. For the simulator kernels to communicate with the PyNN frontend, a intermediate BrainFrame-specific PyNN module (pynn.brainframe) is required that implements and extends common methods and objects like the neuron models, synapse models and projection methods and objects.  In the case of the proposed BrainFrame module, we implemented objects and methods: i) for the initialization of the simulator, ii) for the description of the neuronal network in PyNN, and iii) for controlling the simulation execution. In some cases, an additional interpreter module is needed to translate these Python objects and parameters to each simulator's native parameters and language. For our system, we developed PyHet -- the BrainFrame-specific Python interpreter -- which serves the aforementioned role and also implements the accelerator selection.

The final BrainFrame System will be implementing more generic kernel libraries that will be used by the PyNN front-end to simulate user defined models. That way, the accelerator implementation will be completely transparent to the user and
predictions can then be made based on the analysis of the individual kernels that can guide the selection algorithm.

%Currently, this decision is implemented as an offline check, but we intend to implement a online version of the selection middle-ware to serve experiments in which the problem parameters are updated dynamic during execution.

% ====================================================================
% ====================================================================
\section{Results}

In this section, we present a thorough performance analysis of our heterogeneous BrainFrame platform. The goal is to evaluate the platform and give a clear view on how each accelerator performs when running various instances of the InfOli use cases, validating the usefulness of a heterogeneous HPC simulation framework for computational neuroscience. The performance analysis also acts as a guide for proposing an \emph{accelerator-selection algorithm}.

To validate the correct functionality of the separate accelerator implementations, we use a simple experiment that recreates a typical response that is found in the inferior-olive network (axon response). In this experiment, each cell produces a so-called complex spike, seen in Figure~\ref{io-star}c, from all simulated cells. 6 seconds of brain time are simulated, which translates to 120,000 simulation steps. The complex spike is produced by applying a small current pulse as input to all InfOli cells at the same instance after program onset, for about 500 simulation steps (or 25 ms, in brain time). Despite being rudimentary, this experiment is easy to validate, provided all neurons are initialized with the same state, and also gives a good indication whether synchronization between neurons is correct, thus validating cell interconnectivity (when present).

As mentioned in the introduction, we identify two distinct tracks that can be followed in conducting neuroscientific experiments, both covered in this evaluation. We perform one batch of measurements ranging from 96 to 960 neurons representing small-scale, real-time TYPE-I experiments, and a second batch ranging from 960 to 7,680 neurons representing larger-scale TYPE-II experiments). We consider (by consulting our neuroscience experts) the minimum meaningful network size for experiments to be around 100 neurons, thus our measurements for TYPE-I experiments begin at 96 neurons.
The evaluation is focused the performance of single-node accelerators, thus a network-size cap is set by the smallest maximum network supported by each of the three accelerator fabrics: in this case, the DFE fabric limits network sizes to 7,680 cells.

% ====================================================================
\subsection{Performance Evaluation}

All performance measurements concerning the Xeon Phi have been carried out through the VTune Amplifier XE 2015 profiling and analysis tool by Intel. Timing measurements on the Maia DFE were taken by measuring the DFE-kernel time inlined within the host code using timestamps before and after the kernel call. Since, the host code (in the CPU) is blocking, only the DFE kernel is active during measurements. The time includes the kernel execution (processing and DRAM data-exchange delay) and the activation delay of the FPGA device. This activation takes about 1 ms, which is negligible compared to the overall execution time that takes several seconds to several minutes in our test experiment. GPU kernel-time measurements were taken using the CUDA Event API.

% ====================================================================
\subsubsection{TYPE-I Experiments}

\begin{figure}
    \centering
    \includegraphics[width=0.46\textwidth]{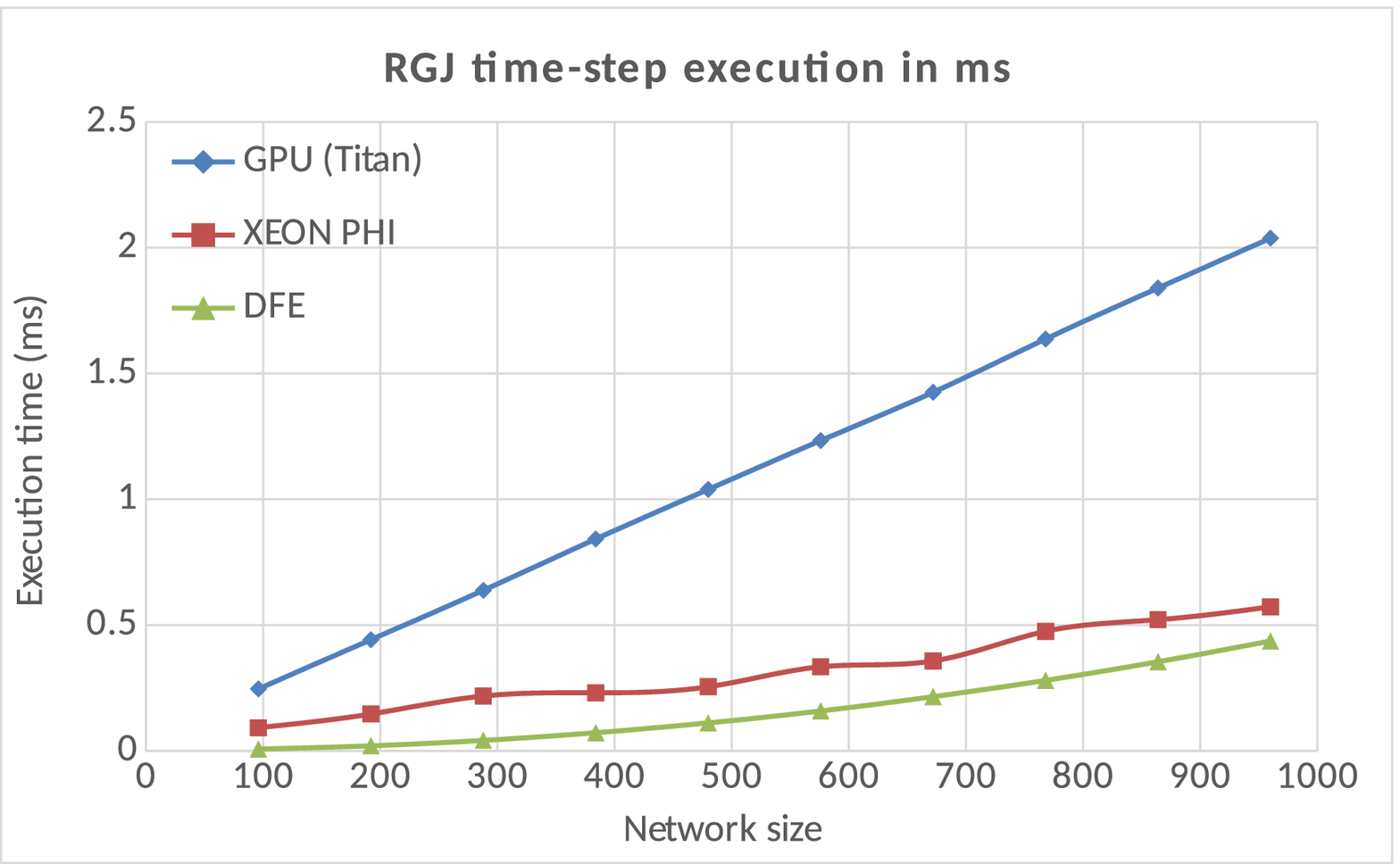}
    \caption{RGJ execution time (TYPE I, 100\% connectivity).}
    \label{rgj100t1}
    \vspace{0.3cm}

    \includegraphics[width=0.46\textwidth]{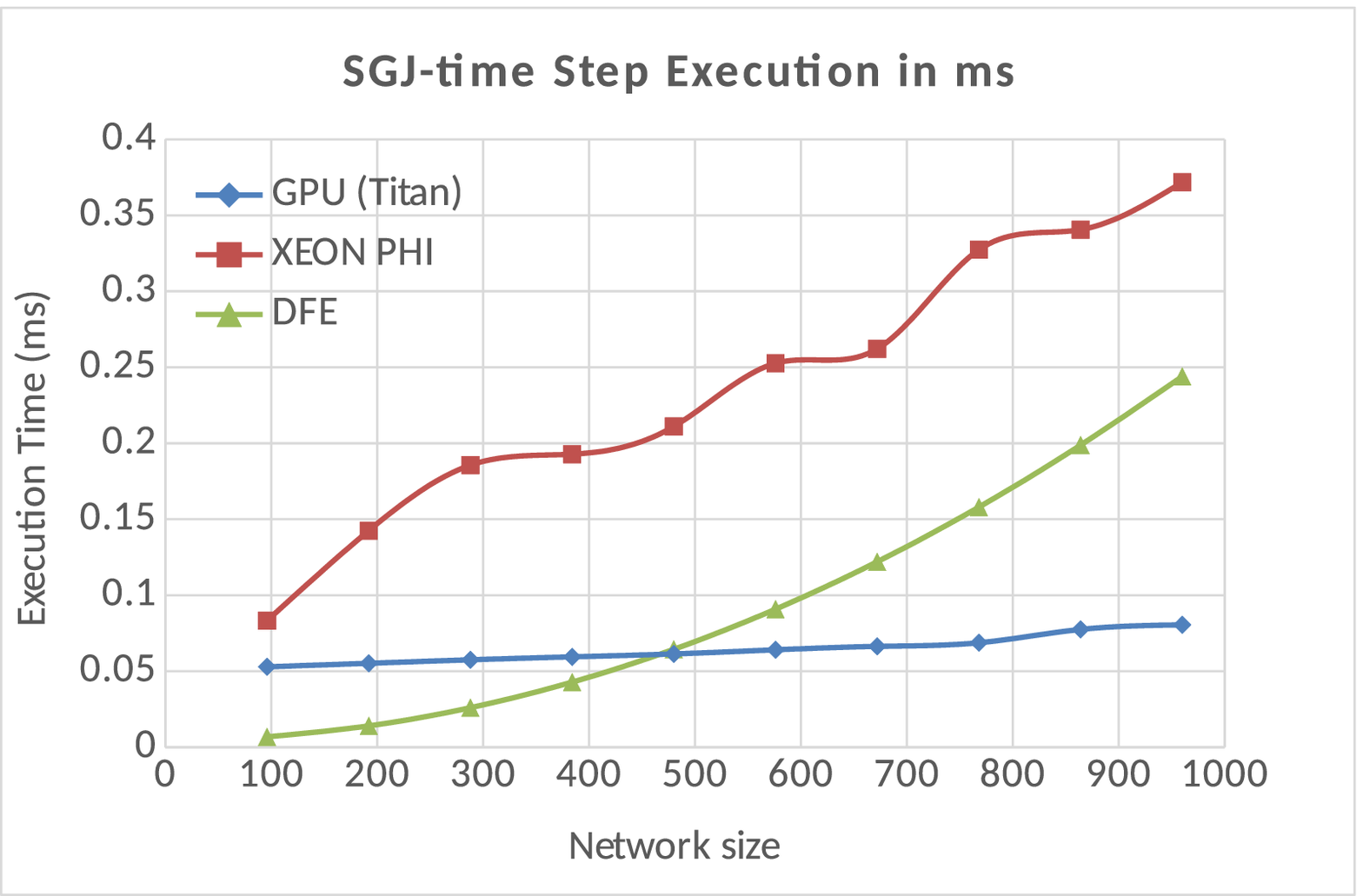}
    \caption{SGJ execution time (TYPE I, 100\% connectivity).}
    \label{sgj100t1}
    \vspace{0.3cm}

    \includegraphics[width=0.46\textwidth]{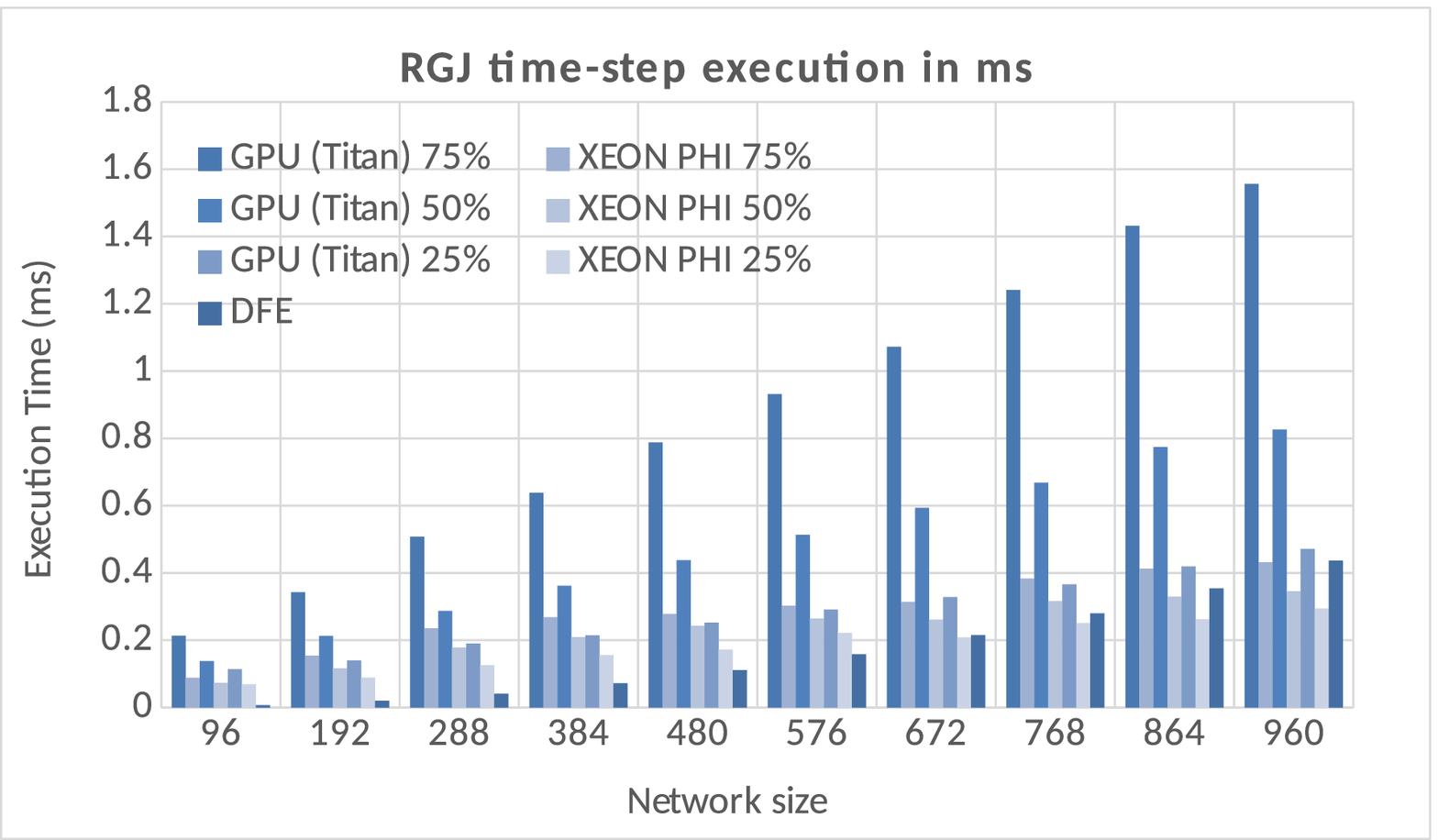}
    \caption{RGJ execution time (TYPE I, $<$100\% connectivity).}
    \label{RgjLess100t1}
    \vspace{0.3cm}

    \includegraphics[width=0.46\textwidth]{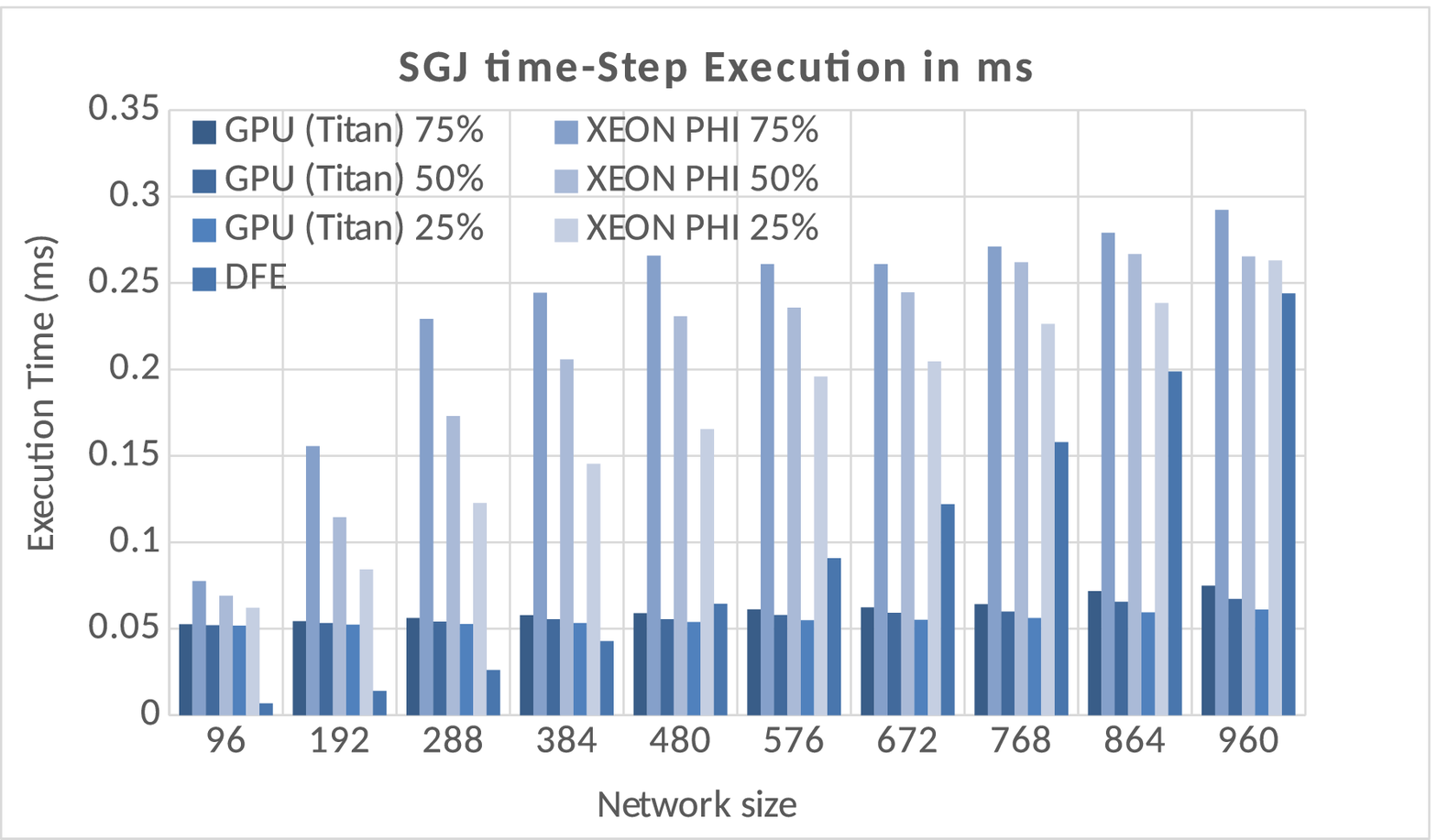}
    \caption{SGJ execution time (TYPE I, $<$100\% connectivity).}
    \label{SgjLess100t1}
\end{figure}

\begin{figure}
    \centering
    \includegraphics[width=0.46\textwidth]{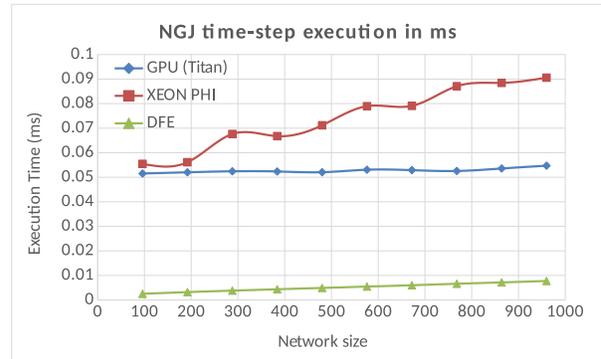}
    \caption{NGJ execution time (TYPE I, no connectivity).}
    \label{Ngjt1}
    \vspace{-0.3cm}
\end{figure}

Starting with the analysis for TYPE-I experimentation, in Figure~\ref{rgj100t1} we plot the execution time of a single simulation time-step ($50~\mu sec$) for the most demanding use case, that of the RGJ with 100\% connectivity density. Even though still not the most common case, a brain-simulation platform must support such high interconnectivity densities for certain TYPE-I experiments. The DFE exhibits the best performance for all tested network sizes. The Xeon Phi is a close second due to the local-memory delays and the less efficient use of its parallel threads: These network sizes are not large enough to provide sufficient parallelism for the Phi threads to be fully utilized. The GPU, on the other hand, has difficulties to cope with the computational intensity of the GJs, which involve mostly division and exponent FP calculations. Since each CUDA thread executes a single neuron, it cannot exploit any potential parallelism in the GJ calculation. This, alongside the fact that the CUDA threads are underutilized at such network sizes, impacts performance drastically.

The inefficiency of the Titan X GPU in performing the realistic GJ computations is clearly revealed in the SGJ case, next (see Figure~\ref{sgj100t1}). In this use case, that the most demanding GJ calculations are dropped, the GPU presents excellent scalability as the problem size increases, compared to the RGJ case. The Xeon Phi, on the other hand, still suffers from core-to-local-memory synchronization delays even though the actual calculations are much simpler now. The DFE needs to spend the same amount of operation ticks as in the RGJ case to evaluate the connection influence, even though it does enjoy gains in performance because of the simpler calculations involved (achieving higher operation frequencies, larger GJ computation parallelism and shorter pipelines). As a result, both latter accelerators show similar scaling properties to the RGJ case. In contrast, the GPU scores performance benefits in the SGJ case compared to the robust DFE for network sizes above 480 neurons.

\begin{table}
    \small
    \begin{center}
    \caption{\label{RT}RT-achievable network size (\#cells) for each use case}
    \item[]\begin{tabular}{l|rrr}
        \br
        \textbf{Use case} & \textbf{~~~~~DFE} &  \textbf{Xeon Phi} &  \textbf{~~~~~GPU}\\
        \br
    	RGJ (100\%)   & 310		& -  	& - \\
        RGJ (75\%)	  & 310 	& - 	& -	\\
        RGJ (50\%)	  & 310     & - 	& -	\\	
        RGJ (25\%) 	  & 310     & - 	& - \\
        \hline
        SGJ (100\%)	  & 400 	& - 	& - \\
        SGJ (75\%)	  & 400   	& - 	& - \\
        SGJ (50\%)	  & 400 	& - 	& 96 \\
    	SGJ (25\%)	  & 400 	& - 	& 96 \\
        \hline
    	NGJ		      & 7,680 	& 96 	& 500 \\
    	\br
    \end{tabular}
    \end{center}
    \vspace{-0.3cm}
\end{table}

Next, it is interesting to evaluate the three accelerators for connectivities of lower than 100\% density. Although not relevant for the DFE which maintains the same implementation for any connectivity density, smaller densities can influence the Xeon Phi and the GPU performance considerably. In Figure~\ref{RgjLess100t1}, we plot the execution time of a single simulation time-step for 25\%, 50\% and 75\% connectivity densities, under the RGJ case. The GPU delivers significant gains but the inefficient GJ execution still causes it to perform worse than DFE, even though the latter operates as in a 100\%-density simulation. The Xeon Phi, on the other hand, manages to achieve enough performance gains to become faster than the DFE for sufficiently large problem sizes; that is, sizes $\ge$960 neurons for 75\% density, $\ge$864 neurons for 50\% density and $\ge$672 neurons for 25\% density.

Under the SGJ use case (Figure~\ref{SgjLess100t1}), we see similar trends as for the 100\% SGJ use case: The GPU exhibits great scalability and is the best option for network sizes higher than 480 neurons. Besides, the DFE remains the most beneficial option for networks smaller than 480.

Under the NGJ case (no connectivity), for TYPE-I experiments, the results point to the DFE as the uniformly best option. In the complete absence of inter-neuron connectivity, the application becomes a purely dataflow workload, fully compatible for acceleration on a DFE, which is tailor-made for such cases, providing significant benefits over both the Xeon Phi and the GPU (see Figure~\ref{Ngjt1}).

Lastly, recall that for TYPE-I experiments, real-time speeds are often desired. Table~\ref{RT} presents the real-time achievable networks for each use case. The results show that, for real-time experimentation, the DFE accelerator is the best option across the board. In contrast, and as mentioned in our previous analysis, the GPU and Xeon-Phi parallel threads tend to be underutilized at such small network sizes, even though most of the delays of using them are present. Thus the DFE -- using fine-grain super-pipelined kernels -- can achieve meaningful network sizes at real-time speeds under all use-case instances, according to the objective set in the introduction ($\ge 100$ cells). For low ($\le50\%$) or zero densities, the GPU and Xeon Phi come close to the real-time objective, yet it is interesting to note that the DFE can even support real-time experimentation for TYPE-II experiments under the NGJ case.

% ====================================================================
\subsubsection{TYPE-II Experiments}

\begin{figure}
    \centering
    \includegraphics[width=0.46\textwidth]{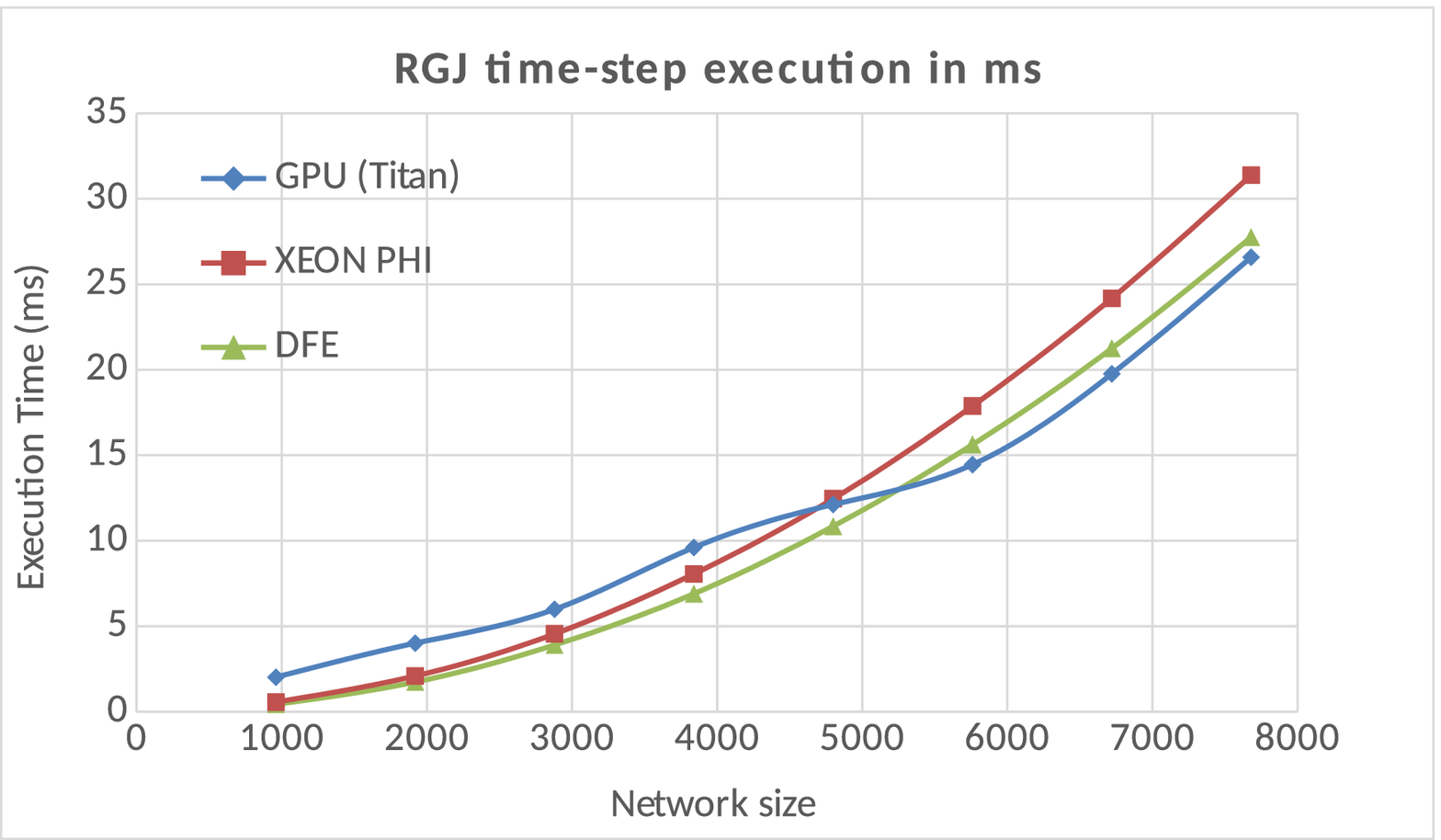}
    \caption{RGJ execution time (TYPE II, 100\% connectivity).}
    \label{Rgj100t2}
    \vspace{0.3cm}

    \includegraphics[width=0.46\textwidth]{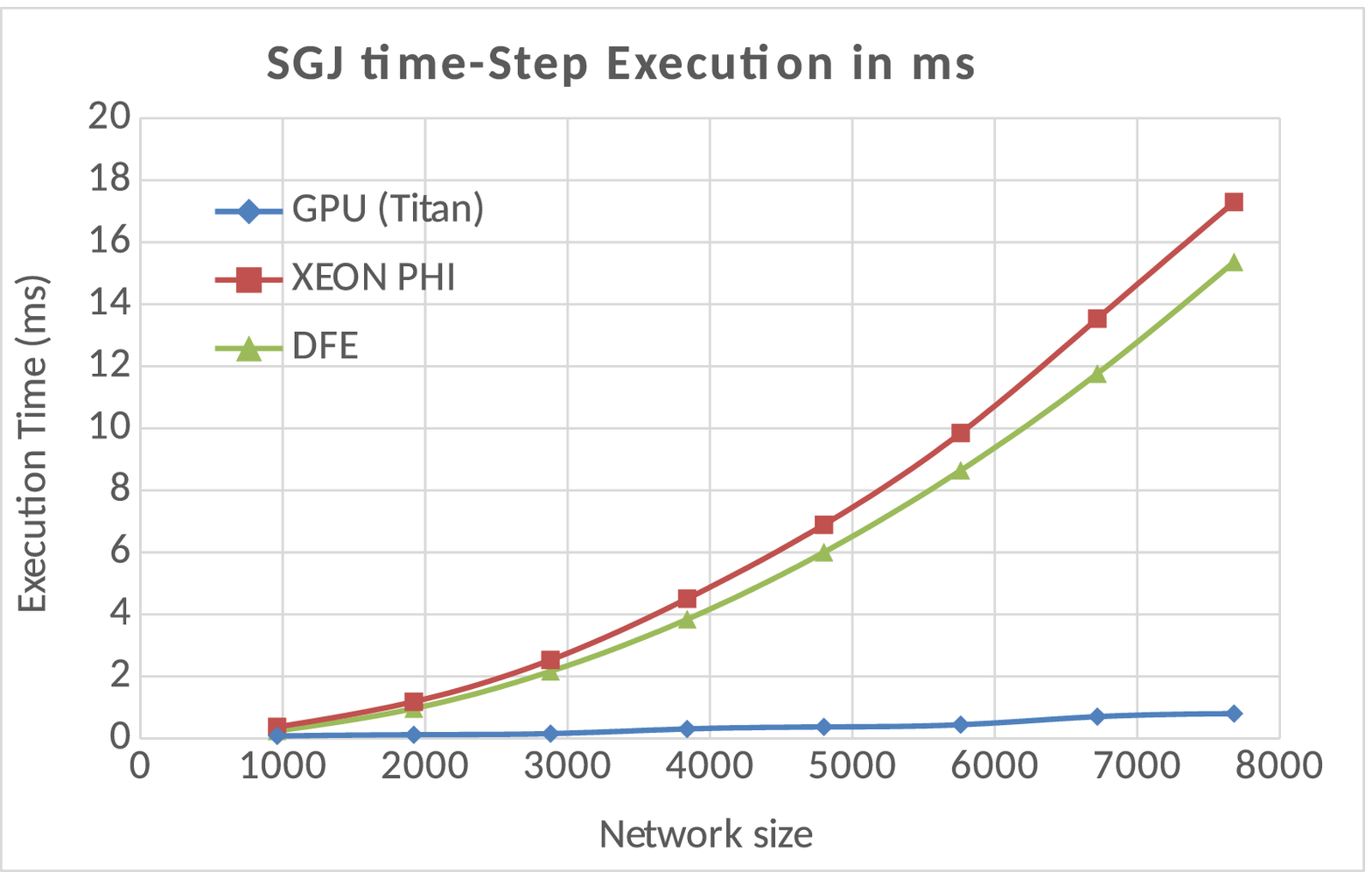}
    \caption{SGJ execution time (TYPE II, 100\% connectivity).}
    \label{Sgj100t2}
    \vspace{0.3cm}

    \includegraphics[width=0.46\textwidth]{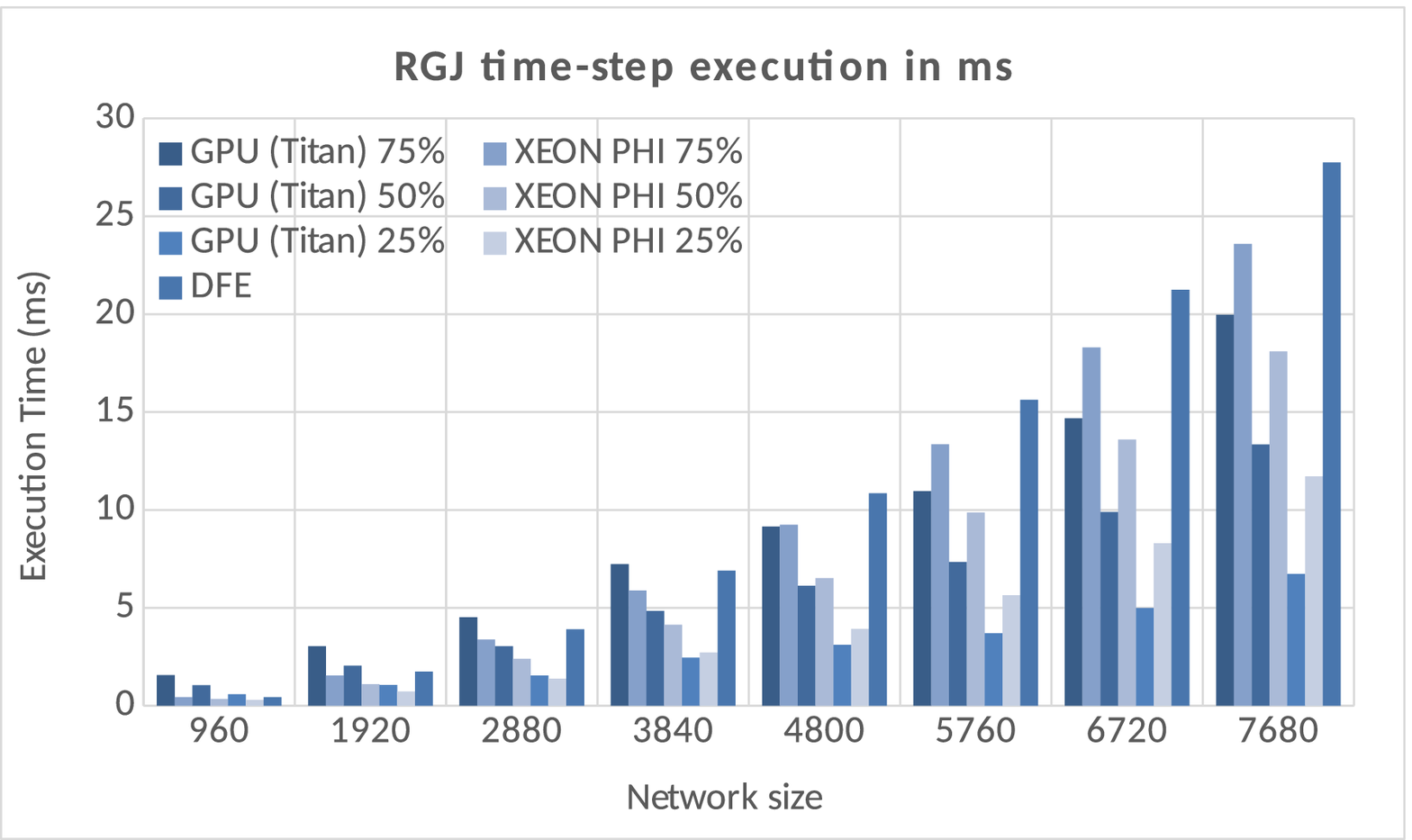}
    \caption{RGJ execution time (TYPE II, $<$100\% connectivity).}
    \label{RgjLess100t2}
    \vspace{0.3cm}

    \includegraphics[width=0.46\textwidth]{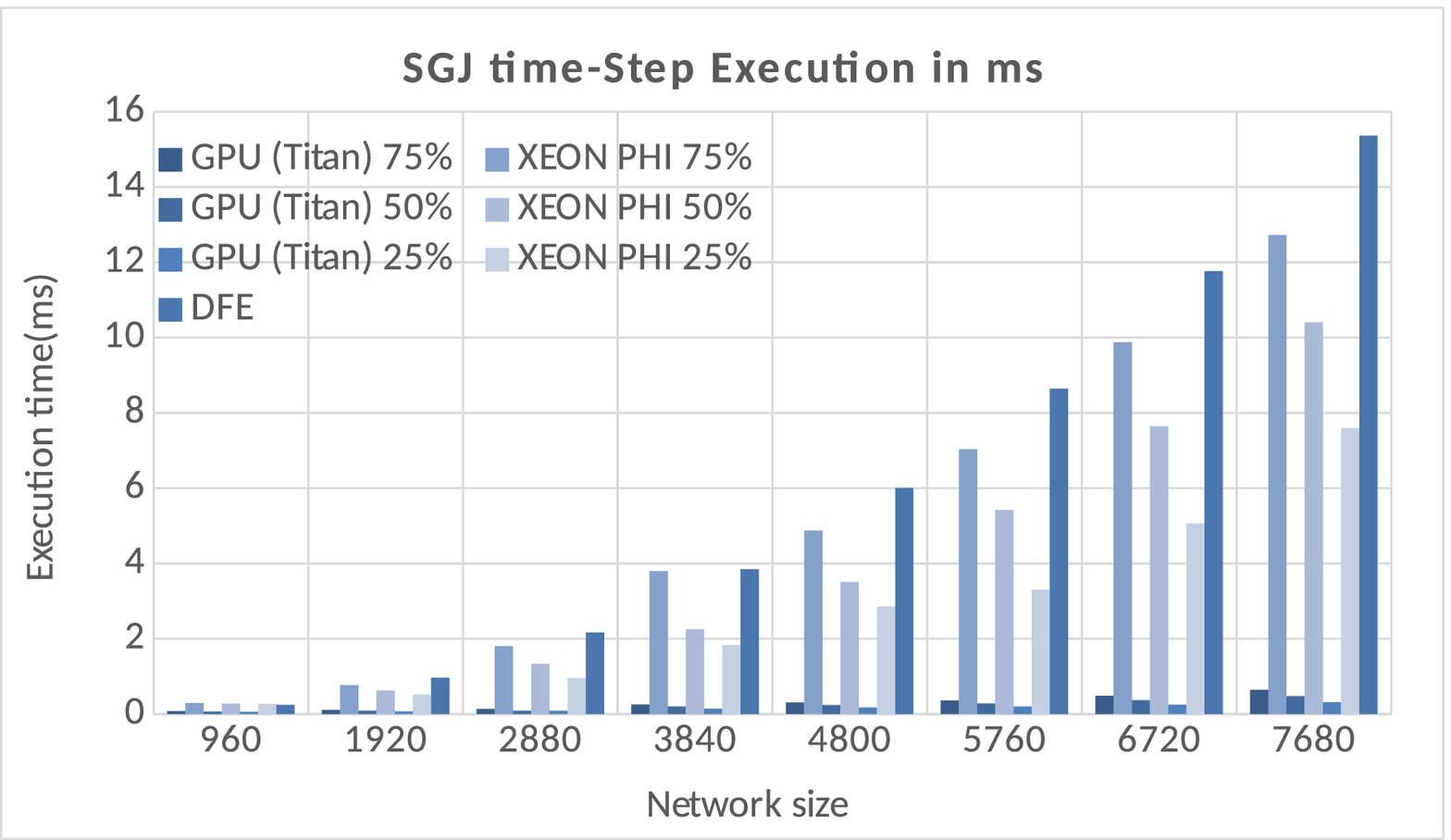}
    \caption{SGJ execution time (TYPE II, $<$100\% connectivity).}
    \label{Sgjless100t2}
\end{figure}

For TYPE-II experiments, the trends under the RGJ case with 100\% connectivity change significantly (see Figure~\ref{Rgj100t2}). Here, the massive explosion of the GJ computations begins to stress the parallelization capabilities of both the Xeon Phi and the DFE. The DFE's efficient parallelization of the GJs relies mostly on its ability to unroll the GJ loop on the FPGA hardware, allowing for more iterations to finish per operation tick. However, the achievable unrolling factor is limited by the available chip area. For network sizes above 1,000 neurons, the DFE compiler is forced to reuse a lot of resources in time (as the unrolling factor is reduced with increasing network sizes). In effect, the dataflow paradigm gradually degenerates to a sequential execution, making the application less scalable on the DFE. The Xeon Phi follows a similar trend, as the communication overhead between cores (which are interconnected through a moderately efficient ring topology~\cite{chatz}) increases, leading to similarly diminished scalability. Opposite to these accelerators, GPU scalability is largely improved. The GPU is underutilized until all CUDA cores are used (3,072) simultaneously, so for experiments over 3,000 neurons scalability is gradually improving. As a result, the GPU becomes the better performing solution (surpassing the DFE) for network sizes of 4,800 neurons and above.

\begin{figure}
    \centering
    \includegraphics[width=0.46\textwidth]{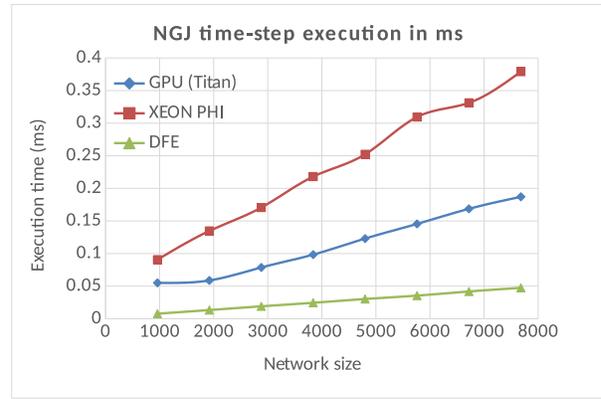}
    \caption{NGJ execution time (TYPE II, no connectivity).}
    \label{Ngjt2}
    \vspace{-0.3cm}
\end{figure}

For lower connectivity densities under the RGJ case, we observe similar trends, although the Xeon-Phi scalability is slightly better because of the lower interconnectivity (see Figure~\ref{RgjLess100t2}). Thus, the Xeon Phi retains the advantages it has for lower than 100\% densities, compared to the DFE. Still, the effect of the inter-core communications is present allowing for the GPU to overtake the Xeon Phi for network sizes above 4,800 neurons (for densities of 50\% and 75\%) and above 3,840 neurons (for 25\% density).

Under the SGJ case, the DFE and Xeon Phi follow similar trends, although they are less pronounced (see Figures~\ref{Sgj100t2} and~\ref{Sgjless100t2}). As in the RGJ case, the GPU maintains its lead over the other two accelerator types for all tested network sizes and connectivity densities. Finally, in the NGJ case, the situation is the same as with TYPE-I experiments: The purely dataflow nature of the application allows the DFE to once more score the best performance across the board (Figure~\ref{Ngjt2}).

% ====================================================================
\subsection{Accelerator-Selection Algorithm}

\begin{figure}
    \centering
    \includegraphics[width=0.49\textwidth]{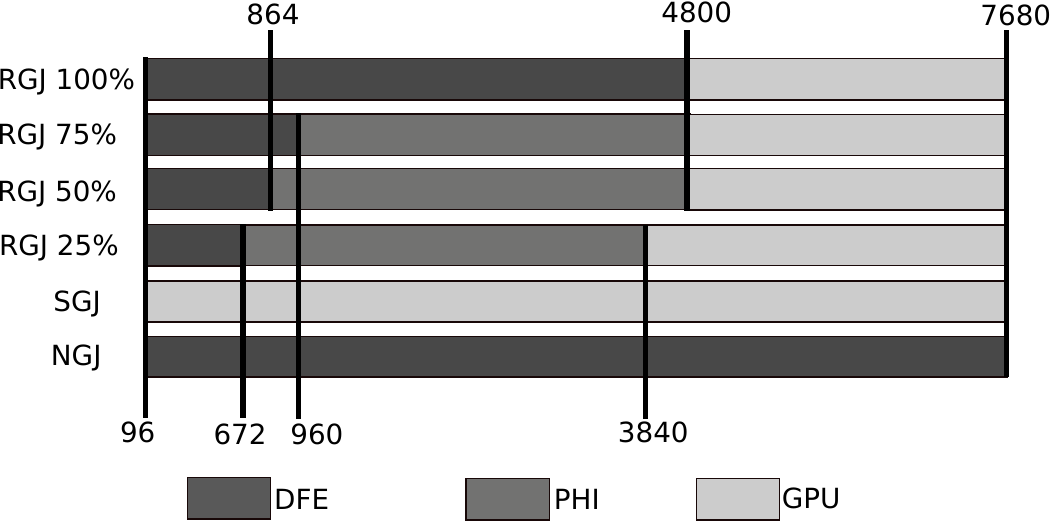}
    \caption{BrainFrame accelerator-selection map for TYPE-II experiments. Selection is heavily dependent on the experiment, involving all three accelerator fabrics. For TYPE-I experiments, the DFE is always the optimal choice (not shown).}
    \label{sel}
\end{figure}

The performance analysis discussed above can now be used to formulate a simple accelerator-selection algorithm for BrainFrame automatically choosing the best-suited accelerator fabric based on the problem parameters: mainly, connectivity detail (biophysically realistic: RGJ, simple: SGJ and not present: NGJ), density and network size. Figure~\ref{sel} shows the selection for our use-case instances. The RGJ case selection, which presents the most complex case in terms of accelerator choice, shifts between all three options depending on the connectivity density. For the SGJ case, the GPU is always the accelerator of choice, while for the NGJ case the DFE yields optimal results under all experiment parameters. Lastly, if the experiment is flagged as a real-time experiment, the algorithm exclusively chooses the DFE to accelerate the application, as it is the only clearly viable accelerator for real-time experiments.

As a simple example of how this selection can speed up experiments, we can assume a scenario where several batches of RGJ experiments need to be executed for various network sizes. Let us assume that each batch includes 5 experiments each with gradually decreasing connectivity density (100\%-75\%-50\%-25\%-0\%) and that each experiment in a batch simulates 40 seconds of brain time. The time saving in this example by using the BrainFrame system compared to homogeneous systems that integrate \emph{only a single} accelerator type can be seen on Table~\ref{saves}.

\begin{table}
    \small
    \centering
    \caption{\label{saves}Time savings (in minutes) with BrainFrame for the assumed experimental scenario compared to three homogeneous-accelerator systems. The \% savings are shown in parenthesis.}
    \setlength\tabcolsep{3pt} % default value: 6pt
    \begin{tabular}{r|rrr}
        \br
                              & \multicolumn{3}{c}{\textbf{BrainFrame vs.}} \\
        \cline{2-4}
        \textbf{Network}      & \textbf{DFE}      & \textbf{Titan X}    &  \textbf{Phi} \\
        \textbf{Size}         & \textbf{-only}    & \textbf{-only}      &  \textbf{-only} \\
        \br
    	384                   & 0.0 (0.0\%)		  & 24.2 (86.2\%)  	        & 8.6 (68.7\%) 	      \\
        960	  	              & 3.2 (13.8\%)	  & 45.8 (69.5\%) 	        & 3.0 (12.8\%)	      \\
        5,760	  	          & 1.9 (43.4\%)	  & 54.5 (27.0\%) 		    & 10.7 (6.8\%)	      \\	
        7,680 	  	          & 591.7 (40.0\%)    & 1.9 (0.2\%) 		    & 246.6 (21.7\%)      \\
        All batches  	      & 707.7 (40.0\%)    & 126.4 (10.7\%) 	        & 268.9 (20.2\%)      \\
	    \br
    \end{tabular}
    \vspace{-0.4cm}
\end{table}

\begin{table}
    \small
    \centering
    \caption{\label{Esaves}Energy savings with BrainFrame for the assumed experimental scenario compared to three homogeneous-accelerator systems. We assume nominal (TDP) power figures (see Table~\ref{specs}).}
    \setlength\tabcolsep{3pt} % default value: 6pt
    \begin{tabular}{r|rrr}
        \br
                              & \multicolumn{3}{c}{\textbf{BrainFrame vs.}} \\
        \cline{2-4}
        \textbf{Network}      & \textbf{DFE}      & \textbf{Titan X}    &  \textbf{Phi} \\
        \textbf{Size}         & \textbf{-only}    & \textbf{-only}      &  \textbf{-only} \\
        \br
    	384                   & 0.0\%	  & 91.4\%  	        & 82.5\% 	      \\
        960	  	              & 38\%	  & 86.4\% 	            & 64.9\%	      \\
        5,760	  	          & 51.3\%	  & 60.9\% 		        & 55.1\%	      \\	
        7,680 	  	          & 23\%      & 20.4\% 		        & 43.8\%      \\
        All batches  	      & 27.3\%    & 32.6\% 	            & 45.9\%      \\
	    \br
    \end{tabular}
    \vspace{-0.4cm}
\end{table}

The BrainFrame system can achieve significant benefits compared to the single-fabric systems that can range up to 86\% execution-time reduction. On average, assuming the total runtime of all batches, the BrainFrame system can achieve 40\% reduction compared to a DFE-only system, a 10.7\% reduction to a GPU-only system and a 20.2\% reduction compared to a Phi-only system.

If we consider the nominal scenario of TDP power consumption we can also present an estimation of the energy benefits of using BrainFrame compared to the single node accelerators for our example. The energy saving for specific batches on the example are between 20.4\% to even about 91.4\%. For the all experimental batches the energy saving is between 27\% to 45.9\%. Reduction in energy consumption can greatly reduce operation and maintenance cost especially within a datacenter environment.

Although these figures will vary based on the particular accelerator instances used for the experiments, they give a rough estimate of the time savings that can be obtained by carefully selecting the accelerators for the various experiments. This selection can be easily extended/updated as new features and more generalized model libraries are added for acceleration (making the selection predictive for general cases) or as each acceleration technology is updated in the future.

% ====================================================================
% ====================================================================
\section{Discussion}

There are numerous related works that propose employing HPC solutions for the acceleration of SNNs. Such solutions include hardware-based solutions, like reconfigurable hardware, as well as software solutions using GPUs and less often many-core processors platforms, such as the Xeon Phi. Simpler modeling has found a good match on GPU-based systems, such as Izhikevich and I\&F modeling~\cite{nageswaran, Yamazaki_NN_2013}. Higher biophysically meaningful modeling, like the extended-HH model, seems to be a much more difficult problem to solve with GPUs, especially for real-time experimentation~\cite{hoang}.

The Xeon Phi has also been used very successfully for bio-inspired neural networks, such as Convolutional Neural Networks for Deep Learning Systems~\cite{Viebke}. On the other hand, similar difficulties to the GPUs in the acceleration of the complex HH models, are identified with Xeon Phi platforms even for less densely interconnected networks~\cite{chatz}.

FPGA-based solutions have been especially prominent in accelerating neuron applications, with impressive results specifically for biophysically meaningful modeling and real-time performance for such networks~\cite{smaragdos2, glackin2, Shayani1}. It is also revealed in related works conducting performance analysis, that an FPGA's potential benefit varies greatly between SNN types, even without taking into account connectivity modeling that can decisively change the workload characteristics~\cite{bhuiyan}.

Recently, we have also seen use of DFEs for accelerating computational-neuroscience applications. On purely dataflow neuromodeling applications, the DFE can have great benefits for both large-scale networks and real-time networks performance~\cite{cheung2}. Even in the cases of HH neurons that include highly accurate interconnectivity modeling (disrupting the purely dataflow nature), the DFEs can accomplish greater benefits than traditional control-flow-based FPGA acceleration~\cite{smaragdos}.

These works, however, present just one-off implementations of specific application instances, on a specific acceleration platform and most also ignore the variety of synapse-modeling alternatives and its influence on the applications. Biophysically accurate models of biological systems, such as the ones using the HH formalism, are comprised mostly of a set of computationally challenging deferential equations often implementing an oscillatory behavior. If neurons are simulated as independent computational islands (NGJ case), then dependencies between the equations do not arise, allowing divide-and-conquer, data-flow and event-driven acceleration strategies to be used very efficiently. The moment interconnectivity between oscillating neurons is also modeled (like GJs, input integrators, STDP synapses etc), the cells become coupled oscillators. The embarrassingly parallel and dataflow nature of the application is then broken. All neuron states need to be completely updated at each simulation step to retain correct functionality. This requirement, in turn, enforces the use of cycle-accurate, transient simulators and forbids event-driven implementations. As a result, a single HPC fabric cannot be a universal solution to the problem since it is unable to cover all the aforementioned requirements efficiently, as our analysis also reveals.

The above difficulties strongly hint on why most of the computational-neuroscience community has so far meticulously avoided employing HH models and multi-compartmental models with complex connections on large problem sizes using conventional computing machines. The eventual use of biophysically plausible neurons and connections on a larger scale is anticipated to contribute substantially in explaining biological behavior. Even though the details of the most important system behaviors of the modeled systems must revolve around very specific characteristics of the networks, thus can possibly to be revealed by generally simpler representations, the computational neuroscientist cannot know beforehand which of the numerous dynamics revealed from the biological measurements (from which the models arise) can be safely abstracted. Studies seeking to reveal system behavior need to start with complex representations before they know enough to drop back to more simplified models.

Additionally, most related works seem to suffer from a limited re-usability value due to their (often inexistent) user interface. They ignore the challenge of the neuroscientific community adopting the proposed platform and very few propose solutions to that end.  Beuler et al.~\cite{beuler} developed a graphical interface alongside their FPGA-based simulator. Although it does provide ease of use in experiments, it is still confined to only one platform and only one application with limited flexibility to be the basis of a more widely adopted system. Weinstein et al.~\cite{weinstein, weinstein_thesis} took the approach of developing their own modeling language to interface to their acceleration library, the DYNAMO compiler. Despite the limitation of using only FPGAs as the back-end platform, the DYNAMO compiler is a technically complete solution. Unfortunately, it failed to achieve wide adoption by the scientific community as it requires learning a new language and, additionally, the non-trivial process of porting all existing neuron models to the new coding paradigm.

PyNN has also been used in the past to tackle the issue of user interface. One such a system is SpiNNaker~\cite{spinnaker}, which uses PyNN to interface to a neuromorphic network comprised by a many-core system based on ARM processors. SpiNNaker though is focused on simpler modeling paradigms that do not model specific biophysical neuron properties and have more tractable computational requirements. The most promising solution, both in terms of usability and computational ability, was proposed by Cheung et al.~\cite{neuroflow} with NeuroFlow. In this work, the researchers integrated PyNN to their DFE-based hardware library. Neuroflow also provides a very complete library of IPs in the back-end, covering a great portion of possible applications. Yet, the system is still integrating a single acceleration platform. What is more, the performance and efficiency analysis is only presented for a single use case of a generally simpler model (Izhikevich) and with connectivity modeling of medium complexity (STDP) and relatively lower density (about 10\%). The behavior and performance of the system for the rest of the supported features is not self-evident and is expected to be significantly different, especially for accurate modeling such as the HH and with high connectivity densities, as shown by our performance analysis on the DFE platform. Furthermore, many of the performance benefits are accomplished using event-driven simulations (neurons are evaluated only when their inputs are triggered), that cannot always be employed, as discussed earlier.

To the best of our knowledge, no prior work has considered an heterogeneous acceleration system for coping with the variability of the applications in the field. Additionally, the PyNN integration provides a familiar interface to the neuroscientific community, thus making BrainFrame a complete solution for a node-level heterogeneous system. Even though the current work introduces BrainFrame in a single-node setup that integrates all three accelerator fabrics using the PCI-e interface, the BrainFrame paradigm is primary designed to support multi-node setups. Such setups can be facilitated in the now up-and-coming heterogeneous datacenters, provided crucial aspects such as low-latency interconnects are tackled. Such a development would lead to a dramatic increase in the size of network populations supported at tractable simulation times, while also providing a way for small-medium-sized labs to use BrainFrame as a service. Thus, enabling them to exploit the benefits of such an HPC platform without suffering the cost of creating and maintaining a local setup.

% ====================================================================
% ====================================================================
\section{Conclusions}

In this paper, we have proposed BrainFrame, an heterogeneous acceleration platform to serve computational-neuroscience studies in conducting the variety of real experimentation often required for the study of brain functionality. We have focused our ana\-lysis on biophysically-accurate neuron models, as such models are considered essential for the deeper understanding of the system properties of biological brain networks.  In order for the BrainFrame system to cope with the demand for high ease of programming use as well as the computational requirements of the field, we have presented a proof-of-concept HPC platform that integrates three accelerator technologies already proven in brain simulations. The performance analysis of the system employing use cases that take into account connectivity density and modeling complexity, has revealed that all three fabrics are essential within such a powerful simulation platform so as to optimally serve all possible experimentation cases. The platform, thus, achieves efficient large-network experiments as well as real-time performance for meaningful network sizes ($\ge 100$ cells).

BrainFrame is complemented, finally, with a PyNN front-end so as to tackle the much sought usability objective. The PyNN front-end makes the heterogeneous platform immediately accessible to a multitude of prior modeling works, which is an essential strategy for the wide adoption of complex HPC platforms in the neuroscientific community. Furthermore, building on the elegant PyNN infrastructure, a simple accelerator-selection algorithm has also been developed for automatically identifying the most suitable HPC fabric (Xeon Phi, GPU, DFE) per neuroscientific experiment and has been integrated in BrainFrame. Last but not least, all accelerators use PCIe slots to connect to the host system, which greatly amplifies the platform flexibility and permits adjusting the platform hardware depending on the funds and hardware resources available to a research lab wishing to use BrainFrame.

% ====================================================================
% ====================================================================
\section{Acknowledgements}

This work is partially supported by the European- Commission Horizon 2020 Framework Programme Project VINEYARD (\emph{Gr. Agr. $N^{o}$ 687628}) and ERC-PoC-2014 project BrainFrame (\emph{Gr. Agr. $N^{o}$ 641000}). We also like to thank the STFC Hartree Centre (UK) for providing the Maxeler and Xeon-Phi computational resources used in our experiments. We gratefully acknowledge the support of NVidia Corporation with the donation of the Titan X GPU used in this research and the continuous support provided by Maxeler Technologies throughout our research effort.

% ====================================================================
% ====================================================================
\section{References}
\bibliographystyle{ieeetr}
\bibliography{ref}

\begin{thebibliography}{10}

\bibitem{jornt}
J.~R. {De Gruijl}, B.~Paolo, G.~de~Jeu Marcel~T., and D.~Z.~C. I., ``{Climbing
  Fiber Burst Size and Olivary Sub-threshold Oscillations in a Network
  Setting},'' {\em PLoS Comput Biol}, vol.~8, 12 2012.

\bibitem{chatz}
G.~Chatzikonstantis, D.~Rodopoulos, S.~Nomikou, C.~Strydis, C.~I. {De Zeeuw},
  and D.~Soudris, ``{First Impressions from Detailed Brain Model Simulations on
  a Xeon/Xeon-Phi Node},'' in {\em {Proceedings of the ACM International
  Conference on Computing Frontiers}}, {CF '16}, (New York, NY, USA),
  pp.~361--364, ACM, 2016.

\bibitem{hoang}
H.~D. Nguyen, Z.~Al-Ars, G.~Smaragdos, and C.~Strydis, ``{Accelerating complex
  brain-model simulations on GPU platforms },'' in {\em {Design, Automation,
  and Test in Europe, DATE 2015}}, Mar. 2015.

\bibitem{smaragdos2}
G.~Smaragdos, S.~Isaza, M.~V. Eijk, I.~Sourdis, and C.~Strydis, ``{FPGA-based
  Biophysically-Meaningful Modeling of Olivocerebellar Neurons},'' in {\em
  {22nd ACM/SIGDA International Symposium on Field-Programmable Gate Arrays
  (FPGA)}}, Feb. 2014.

\bibitem{glackin2}
B.~Glackin, J.~A. Wall, T.~M. McGinnity, L.~P. Maguire, and L.~McDaid, ``{A
  spiking neural network model of the medial superior olive using spike timing
  dependent plasticity for sound localization},'' {\em Frontiers on Comput.
  Neurosci.}, vol.~4, no.~18, 2010.

\bibitem{bhuiyan}
M.~Bhuiyan, A.~Nallamuthu, M.~Smith, and V.~Pallipuram, ``{Optimization and
  performance study of large-scale biological networks for reconfigurable
  computing},'' in {\em {Fourth International Workshop on High-Performance
  Reconfigurable Computing Technology and Applications ( HPRCTA)}}, pp.~1--9,
  nov. 2010.

\bibitem{Yamazaki_NN_2013}
T.~Yamazaki and J.~Igarashi, ``{Realtime cerebellum: A large-scale spiking
  network model of the cerebellum that runs in realtime using a graphics
  processing unit},'' {\em Neural Networks}, vol.~47, pp.~103--111, 2013.
\newblock Computation in the Cerebellum.

\bibitem{Smaragdos2016}
G.~Smaragdos, G.~Chatzikostantis, S.~Nomikou, D.~Rodopoulos, I.~Sourdis,
  D.~Soudris, C.~I. de~Zeeuw, and C.~Strydis, ``{Performance Analysis of
  Accelerated Biophysically-Meaningful Neuron Simulations},'' in {\em {2016
  Ieee International Symposium on Performance Analysis of Systems and Software
  Ispass 2016}}, pp.~1--11, 2016.

\bibitem{Markram}
H.~Markram {\em et~al.}, ``{Reconstruction and Simulation of Neocortical
  Microcircuitry},'' {\em Cell}, vol.~163, no.~2, pp.~456--492, 2015.

\bibitem{HH}
A.~L. Hodgkin and A.~F. Huxley, ``{quantitative description of membrane current
  and application to conduction and excitation in nerve},'' {\em Journal
  Physiology}, vol.~117, pp.~500--544, 1954.

\bibitem{pynn}
A.~Davison, D.~Br{\"u}derle, J.~Eppler, J.~Kremkow, E.~Muller, D.~Pecevski,
  L.~Perrinet, and P.~Yger, ``{PyNN: a common interface for neuronal network
  simulators},'' {\em Front. Neuroinform}, vol.~2, no.~11, 2008.

\bibitem{DeGruijl8937}
J.~R. {De Gruijl}, T.~M. Hoogland, and C.~I. {De Zeeuw}, ``{Behavioral
  Correlates of Complex Spike Synchrony in Cerebellar Microzones},'' {\em
  Journal of Neuroscience}, vol.~34, no.~27, pp.~8937--8947, 2014.

\bibitem{Hoogland}
T.~M. Hoogland, J.~R.~D. Gruijl, L.~Witter, C.~B. Canto, and C.~I.~D. Zeeuw,
  ``{Role of Synchronous Activation of Cerebellar Purkinje Cell Ensembles in
  Multi-joint Movement Control},'' {\em Current Biology}, vol.~25, no.~9,
  pp.~1157--1165, 2015.

\bibitem{dezeeuw}
{C.I. De Zeeuw}, {F.E. Hoebeek }, {L.W.J. Bosman}, {M. Schonewille}, {L.
  Witter}, and {S.K. Koekkoek}, ``{Spatiotemporal firing patterns in the
  cerebellum},'' {\em Nat Rev Neurosci}, vol.~12, pp.~327--344, jun 2011.

\bibitem{gao}
Z.~Gao, B.~J. van Beugen, and C.~I.~D. Zeeuw, ``{Distributed synergistic
  plasticity and cerebellar learning},'' {\em Nature Reviews Neuroscience},
  vol.~13, pp.~619--635, Sept. 2012.

\bibitem{profiler}
C.~Feenstra, ``{A Memory Access and Operator Usage Profiler Framework for HLS
  Optimization: Using the Lucas Optical Flow Algorithm as Case Study},''
  Master's thesis, EEMCS, Circuits and Systems, TuDelft, 2011.

\bibitem{Pell2013}
O.~Pell, O.~Mencer, K.~H. Tsoi, and W.~Luk, {\em {Maximum Performance Computing
  with Dataflow Engines}}, pp.~747--774.
\newblock New York, NY: Springer New York, 2013.

\bibitem{jeffers2013intel}
J.~James and J.~Reinders, {\em {Intel Xeon Phi coprocessor high-performance
  programming.}}
\newblock 2013.

\bibitem{nvidia}
{NVidia Corporation}, ``{ www.geforce.com}.''

\bibitem{smaragdos}
G.~Smaragdos, C.~Davies, C.~Strydis, I.~Sourdis, C.~Ciobanu, O.~Mencer, and
  C.~{De Zeeuw}, ``{Real-Time Olivary Neuron Simulations on Dataflow Computing
  Machines},'' in {\em {Supercomputing}} (J.~Kunkel, T.~Ludwig, and H.~Meuer,
  eds.), vol.~8488 of {\em {Lecture Notes in Computer Science}}, pp.~487--497,
  Springer International Publishing, 2014.

\bibitem{nageswaran}
J.~Nageswaran, N.~Dutt, J.~Krichmar, A.~Nicolau, and A.~Veidenbaum,
  ``{Efficient simulation of large-scale spiking neural networks using CUDA
  graphics processors},'' in {\em {Neural Networks, 2009. IJCNN 2009.
  International Joint Conference on}}, pp.~2145--2152, IEEE, 2009.

\bibitem{Viebke}
A.~Viebke and S.~Pllana, ``{The Potential of the Intel Xeon Phi for Supervised
  Deep Learning },'' in {\em {17th IEEE International Conference on High
  Performance Computing and Communications (HPCC 2015)}}, June 2015.

\bibitem{Shayani1}
H.~Shayani, P.~Bentley, and A.~M. Tyrrell, ``{Hardware Implementation of a
  Bio-plausible Neuron Model for Evolution and Growth of Spiking Neural
  Networks on FPGA},'' in {\em {NASA/ESA Conf. on Adaptive Hardware and
  Systems}}, pp.~236--243, June 2008.

\bibitem{cheung2}
K.~Cheung, S.~R. Schultz, and W.~Luk, ``{A large-scale spiking neural network
  accelerator for FPGA systems},'' in {\em {Int. conf. on Artificial Neural
  Networks and Machine Learning}}, {ICANN'12}, pp.~113--120, 2012.

\bibitem{beuler}
M.~Beuler, A.~Tchaptchet, W.~Bonath, S.~Postnova, and H.~A. Braun, ``{Real-Time
  Simulations of Synchronization in a Conductance-Based Neuronal Network with a
  Digital FPGA Hardware-Core},'' in {\em {Artificial Neural Networks and
  Machine Learning -- ICANN 2012}}, September 2012.

\bibitem{weinstein}
R.~K. Weinstein and R.~H. Lee, ``{Architectures for high-performance FPGA
  implementations of neural models},'' {\em Journal of Neural Engineering},
  vol.~3, no.~1, p.~21, 2006.

\bibitem{weinstein_thesis}
R.~K. Weinstein, {\em {Techniques for FPGA neural modeling}}.
\newblock PhD thesis, 2006.

\bibitem{spinnaker}
E.~Painkras, L.~A. Plana, J.~Garside, S.~Temple, S.~Davidson, J.~Pepper,
  D.~Clark, C.~Patterson, and S.~Furber, ``{SpiNNaker: A multi-core
  System-on-Chip for massively-parallel neural net simulation},'' in {\em
  {Proceedings of the IEEE 2012 Custom Integrated Circuits Conference}},
  pp.~1--4, Sept 2012.

\bibitem{neuroflow}
K.~Cheung, S.~R. Schultz, and W.~Luk, ``{NeuroFlow: A General Purpose Spiking
  Neural Network Simulation Platform using Customizable Processors},'' {\em
  Frontiers in Neuroscience}, vol.~9, p.~516, 2016.

\end{thebibliography}

\end{document}